\documentclass[runningheads]{llncs}

\usepackage[final,year=2024,ID=9082]{eccv}

\usepackage{eccvabbrv}

\usepackage{graphicx}
\usepackage{booktabs}

\usepackage[accsupp]{axessibility}  

\usepackage{url}
\usepackage{booktabs}       
\usepackage{amsfonts}       
\usepackage{nicefrac}       
\usepackage{microtype}      
\usepackage{xcolor}         
\usepackage{amsmath}
\usepackage{bbm}
\usepackage{ulem}
\usepackage{graphicx}       
\usepackage{subcaption}
\usepackage{wrapfig}
\usepackage{makecell}
\usepackage{float}
\usepackage{appendix}
\usepackage{algorithm}
\usepackage{algorithmicx}
\usepackage{diagbox}
\usepackage{multirow}

\usepackage[pagebackref,breaklinks,colorlinks]{hyperref}

\usepackage{orcidlink}

\begin{document}

\title{TP2O: Creative Text Pair-to-Object Generation using Balance Swap-Sampling} 

\titlerunning{Creative Text Pair-to-Object Generation}

\author{Jun Li$^{\star\dagger}$\orcidlink{0000-0003-3716-671X} \and
Zedong Zhang$^{\star}$\orcidlink{0000-0002-3328-1713} \and
Jian Yang}

\authorrunning{Jun Li, Zedong Zhang and Jian Yang}

\institute{School of Computer Science and Engineering, \\ Nanjing University of Science and Technology, Nanjing, China\\
\email{\{junli, zandyz, csjyang\}@njust.edu.cn} \\ 
$\star$ equal contribution, $\dagger$ corresponding author}

\maketitle
\begin{figure}[h]
\vskip -0.2in
\hsize=\textwidth
    \centering
\includegraphics[width=0.95\linewidth]{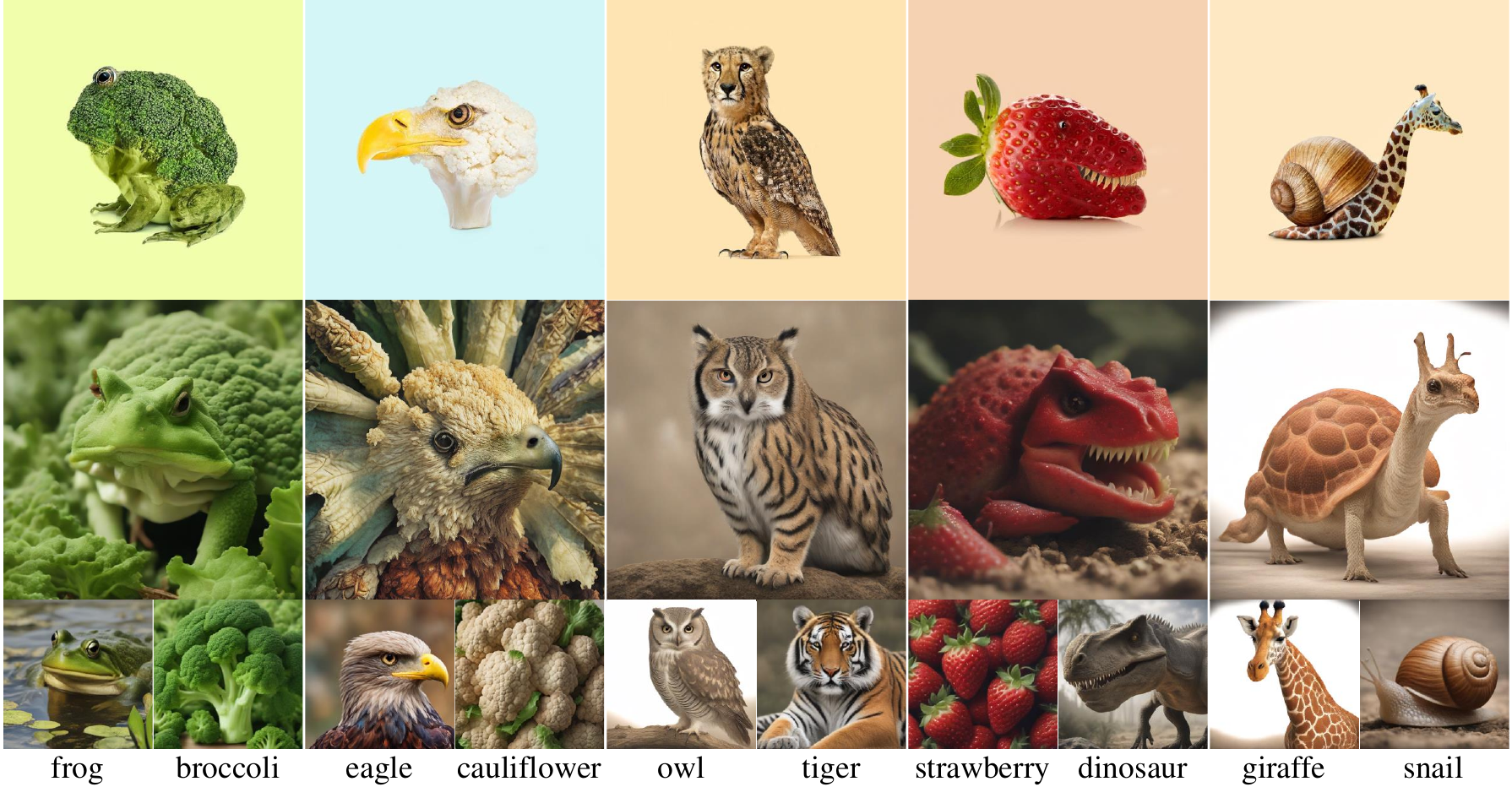}
\caption{ We propose a simple yet effective sampling method without any training to generate creative combinations from two object texts. Bottom row: original images from Stable-Diffusion2 \cite{Rombach2022latentDM}. Middle row: combinations produced by our algorithm. Top row: artworks by \href{https://www.lescreatonautes.fr/}{\textit{Les Créatonautes}}, a French creative agency, from the \href{https://www.instagram.com/les.creatonautes/}{Instagram}. } 

\label{fig:creativeobject}
\vskip -0.3in
\end{figure}

\begin{abstract}
Generating creative combinatorial objects from two seemingly unrelated object texts is a challenging task in text-to-image synthesis, often hindered by a focus on emulating existing data distributions. In this paper, we develop a straightforward yet highly effective method, called \textbf{balance swap-sampling}. First, we propose a swapping mechanism that generates a novel combinatorial object image set by randomly exchanging intrinsic elements of two text embeddings through a cutting-edge diffusion model. Second, we introduce a balance swapping region to efficiently sample a small subset from the newly generated image set by balancing CLIP distances between the new images and their original generations, increasing the likelihood of accepting the high-quality combinations. Last, we employ a segmentation method to compare CLIP distances among the segmented components, ultimately selecting the most promising object from the sampled subset. Extensive experiments demonstrate that our approach outperforms recent SOTA T2I methods. Surprisingly, our results even rival those of human artists, such as \textit{frog-broccoli} in Figure \ref{fig:creativeobject}. \href{https://njustzandyz.github.io/tp2o/}{Project}    
\end{abstract}

\section{Introduction}
\label{sec:intro}
Human creativity plays a crucial role in the innovative visual generation from textual concepts, known as text-to-image (T2I) synthesis, due to its the ability to come up with visual generations that are novel, surprising and valuable \cite{boden2004creative}, such as the combinatorial artworks created by artists in the top of Figure \ref{fig:creativeobject}. However, this task poses a significant challenge for most existing methods in computer vision, including DALLE3 \cite{Ramesh2022DALLE2}, Stable-Diffusion2 \cite{Rombach2022latentDM}, and ERNIE-ViLG2 \cite{Feng2022ERNIE-ViLG2}. These methods aim to generate images that emulate a given training distribution \cite{Elgammal2017creativeAN}, but they often lack the potential for the creativity \cite{Das2022creativity}. Consequently, there is a need to develop a vision system with enhanced creative capabilities.

Recent efforts have primarily focused on compositional objects, aiming to directly generate new and intricate images by composing textual descriptions of multiple known objects. For example, Composable Diffusion Models (CDMs) \cite{liu2022compositional} generates images containing multiple objects at specified positions, and Custom-Diffusion \cite{kumari2023customdiffusion} enables the generation of new and reasonable compositions of multi-objects in previously unseen contexts. However, these methods only produce generations with independent objects, lacking the element of combinatorial creativity (see Fig. 1 \cite{Feng2023trainingfreeT2I} and Fig. 1 \cite{kumari2023customdiffusion}). This raises an interesting task, \textit{Creative Text Pair-to-Object Generation:} produce a creative object using an object text pair, akin to the human artworks using  \textit{frog-broccoli} and \textit{turtle-cat}.

To address this question, we adhere to three essential criteria, \textit{novelty}, \textit{surprise} and \textit{value} \cite{boden2004creative,maher2010evaluating}, for evaluating the creativity of the combinatorial object in the T2I synthesis. Creative combinations often lies in their unpredictability for several reasons. Firstly, these three criteria encompass diverse interpretations, including `psychological' and `historical' novelties, along with unfamiliar, unaware and impossible surprises \cite{boden2004creative}. Moreover, values, particularly in the realm of science, can be elusive and subject to change \cite{boden2004creative}. Secondly, when combinations become predictable, the element of surprise diminishes as they inherently contain less information entropy.
Since T2I models typically rely on datasets with similar distributions, we revisit these criteria when combining two object texts.
\textbf{\textit{Balanced novelty}} quantifies a balance degree between the two object texts, ensuring that the combined object diverges from both. \textbf{\textit{Associated surprise}} measures the improbable association between the two object texts, while \textbf{\textit{human-preference value}} assesses the human appeal of the combined object.

Based on the aforementioned criteria, we propose a novel technique called \textbf{\textit{BAlance Swap-Sampling}} (BASS) that generates combinatorially creative objects by fusing the prompt embeddings of two object concepts. Our approach comprises a text encoder, a swapping mechanism, an image generator, and a balance region using a CLIP metric \cite{radford2021learning}. To begin, the text encoder and image generator can be pretrained using state-of-the-art T2I models, such as Stable-Diffusion2 \cite{Rombach2022latentDM}, and we select impossibly associated categories to form text concept pairs from the ImageNet \cite{ILSVRC15} for implementing their possibly \textbf{\textbf{surprising}} combinations. Initially, we obtain two original embeddings by inputting the prompts of object text pairs into the text encoder, respectively. Then, two original images are generated by using these embeddings in the image generator. Next, we propose a swapping mechanism that interchange intrinsic elements of the prompt embeddings, allowing the image generator to create \textbf{\textit{novel}} combinations. Furthermore, we establish a \textbf{\textit{balance sampling}} region within which the newly created image likely exhibits a high-quality fusion of the two original concepts. This is achieved by carefully managing the CLIP distances between the newly creations and the two originally generations. The balance region allows us to efficiently sample a small subset of the newly created images from a pool of randomly exchanging column vectors. Lastly, we further balance the internal components by employing Segment Anything Model (SAM) \cite{Kirillov2023segany} to compare the CLIP distances among the segmented components, ultimately selecting the most promising combinatorial image from the sampled subset. Surprisingly, when we utilize PickScore \cite{Kirstain2023PickaPicAO} and HPSv2 \cite{wu2023hpsv2}, both trained with human preference datasets, instead of our sampling process, Table \ref{tab:Quantitativeevaluations} reveals that our approach, despite lacking specific training and human preference feedback, still exhibits comparable performance to these models. This underscores that our combined objects hold significant \textbf{\textit{value}} in terms of human preference. Overall, our contributions are summarized as follows:
\begin{itemize}
\item  We explore a balance swap-sampling approach to produce novel and surprising combinatorial objects without any training, based solely on two object texts. To the best of our knowledge, we are the pioneers in developing a visual system with creative combination capabilities in text-to-image synthesis. 
\item Our approach involves swapping intrinsic elements within prompt embeddings to generate novel combinations, while controlling CLIP distances between the resulting combination image and the original images to ensure balance for better founding their valuable combination.  
\item Experimental results demonstrate the effectiveness of our approach in generating previously unseen and creative object images in Figures \ref{fig:creativeobject} and \ref{fig:result_compare}. Our method surpasses the object images generated by the SOTA T2I techniques.
\end{itemize}
\section{Related Work}
Here we mainly review compositional T2I, creativity, and OOD generation.

\textbf{Text-to-image (T2I) synthesis} has garnered increasing attention in recent years due to its remarkable progress \cite{zhu2019dm, Liao_2022_text2image, yu2022scaling}. Notably, diffusion models \cite{ho2020denoising, Kawar2022DDRM, liu2022pseudo, zhao2023unipc, song2023consistency} combined with CLIP models \cite{radford2021learning} have shown great promise in T2I synthesis. For instance, CLIPDraw \cite{Frans2022CLIPDraw} employs only the CLIP embedding to generate sketch drawings. DALLE2 \cite{Ramesh2022DALLE2} introduces a diffusion decoder to generate images based on text concepts. Stable-Diffusion \cite{Rombach2022latentDM} has emerged as the most popular choice due to its open-source nature and ability to save inference time.

\textbf{Compositional T2I}  primarily focuses on generating new and complex images by combining multiple known concepts. These concepts can be composed in various ways, including object-object, object-color and shape, object conjunction and negations, object relations, attributes, and modifying sentences by words. Some notable approaches in this field include object-object compositions \cite{liu2022compositional, Feng2023trainingfreeT2I, kumari2023customdiffusion}, image-concept compositions such as subject-context, segmentation-text, and sketch-sentence \cite{Park2021ct2i, Du2020cvg, liu2022compositional, Chefer2023Attend-and-Excite, Li_2022_StyleT2I, Cong2023Attribute-Centric, Gal2023personalizeT2I, Hertz2023prompt, Orgad2023editing, Ruiz2022DreamBooth, brooks2023instructpix2pix, avrahami2023spatext}. However, a common limitation of these compositional T2I models is that they often generate images based solely on the compositional text descriptions, which restricts their creativity. Moreover, the existing object-object composition approaches \cite{liu2022compositional, Feng2023trainingfreeT2I, kumari2023customdiffusion} tend to produce images with independent objects. Recent Magicmix \cite{Liew2022Magicmix} closely related to ours employs linear interpolation to merge distinct semantic images and text, aiming to produce novel conceptual images. But the outcomes may sometimes exhibit an unnatural blend and lack artistic value. In contrast, we propose an innovative sampling method that enables effective information exchange between two object concepts, leading to the creation of captivating composite images.

\textbf{Creativity}  encompasses the ability to generate ideas or artifacts across various domains, including concepts, compositions, scientific theories, cookery recipes, and more \cite{boden2004creative,Boden1998creativityAI,maher2010evaluating,Cetinic2022artAI,Hitsuwari2023creativeart,wang2023creativebirds,dai2024harmonious}. Recently, there has been significant research exploring the integration of creativity into GANs \cite{Goodfellow2014_GAN,Ge2022CreativeSketch} and VAEs \cite{Kingma2014_VAE}. For instance, CAN \cite{Elgammal2017creativeAN} extend the capabilities of GANs to produce artistic images by maximizing deviations from established styles while minimizing deviations from the art distribution. CreativeGAN, systematically modifies GAN models to synthesize novel engineering designs \cite{nobari2021creativegan,Nobari2021rang-GAN}. Additionally, CreativeDecoder \cite{Das2020creativeDecoder,Cintas2022creativechara} enhances the decoder of VAEs by utilizing sampling, clustering, and selection strategies to capture neuronal activation patterns. 
Recent works \cite{Boutin2022divvsrec,Boutin2023diffartist} design a method to assess one-shot generative models in approximating human-produced data by examining the trade-off between \textit{recognizability} and \textit{diversity} (measured by standard deviation). Unlike these methods, our approach introduces a swapping mechanism to to enhance the generation of novel combinational object images, along with a defined region for accepting high-quality combinations.

\textbf{Out-of-Distribution (OOD)} is closely related to our work as we generate creative object images that lie outside the data distribution. However, existing OOD techniques primarily concentrate on detection tasks \cite{shen2021towards,ye2021ood} through disentangled representation learning \cite{trauble2021disentangled}, causal representation learning \cite{shen2022weakly, khemakhem2020variational}, domain generalization \cite{zhou2020deep,zhou2020learning}, and stable learning \cite{xu2022theoretical}. In contrast, our focus is on OOD generation, where we aim to create meaningful object images. While \cite{Ren2023OOD} also incorporate an OOD generation step for the ODD detection task, they primarily address scenarios where the input distribution has shifted. To augment the limited dataset for the image classification task, diffusion models are employed to produce numerous images through outlier \cite{Du2023outlier} and guidance \cite{Zhang2023gif} imaginations within the same category. In our approach, we create a novel distribution by blending the embedding distributions of two category concepts.

\begin{figure}[t]
\centering
\includegraphics[width=0.97\linewidth]{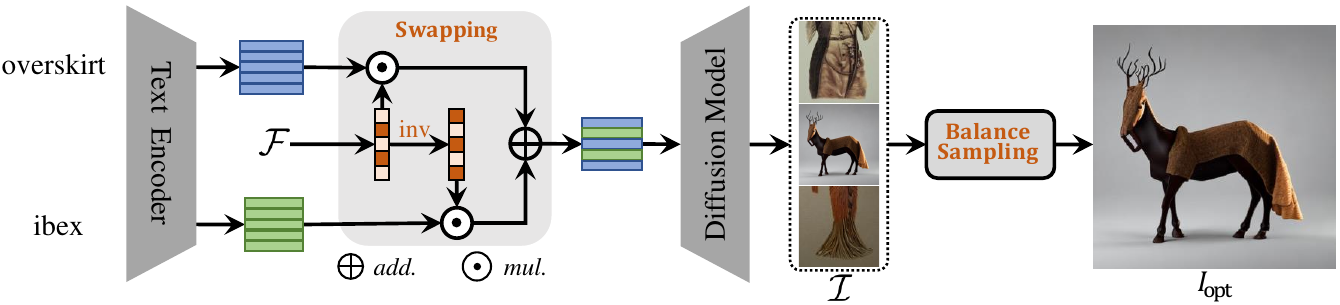}
\caption{The pipeline of our balance swap-sampling method. Starting from text embeddings by inputting two given texts into the text encoder, we introduce a swapping operation to collect a set $\mathcal{F}$ of randomly swapping vectors for novel embeddings, then generate a new image set $\mathcal{I}$, and propose a balance region to build a sampling method for selecting an optimal combinatorial object image.} 
\label{fig:framework}
\end{figure}

\section{Methodology}
In this section, we introduce a balance swap-sampling (BASS) approach for generating a new and surprising object using two text concepts, as shown in Figure \ref{fig:framework}. Its key parts include a swapping mechanism, a balance region and Our BASS method. Before showing the details of our method, we provide an overview of the unified generation process using the T2I models (\eg, Stable-Diffusion2 \cite{Rombach2022latentDM}, DALLE2 \cite{Ramesh2022DALLE2}, and Imagic \cite{Saharia2022text2image}). 

\textbf{T2I:} For a text $t$ and its associated prompt $p$, a generated image is described as $G=\mathcal{G}(E)$, where $E=\mathcal{E}(p)\in \mathbb{R}^{h\times w}$ is a text encoder with dimensions $h$ and $w$, and $\mathcal{G}(\cdot)\in \mathbb{R}^{H\times W}$ is an image generator with dimensions $H$ and $W$. The variable $h$ denotes the maximum number of input words, while $w$ represents the dimensionality of each word embedding. $H$ and $W$ denote the size of the generated image. Note that the text encoder $\mathcal{E}(\cdot)$ and the image generator $\mathcal{G}(\cdot)$ can be pretrained using Stable-Diffusion2 \cite{Rombach2022latentDM}, as our baseline. Alternatively, any other diffusion model can also be utilized in our approach. 

Given a text pair $(t_1,t_2)$, we use the T2I model to generate their original images $I_1=\mathcal{G}(\mathcal{E}(p_1))$ and $I_2=\mathcal{G}(\mathcal{E}(p_2))$, where $p_1$ and $p_2$ are the prompts of $t_1$ and $t_2$, respectively. For example, for a text pair $(\text{\textit{frog}},\text{\textit{broccoli}})$, we use its prompt pair $\text{\textit{(A photo of frog}}, \text{\textit{A photo of broccoli)}}$, to produce two images, please refer to the two below figures in the first column of the Figure \ref{fig:creativeobject}.

\subsection{Swapping Mechanism}
\label{sec:swapping}
Following the generation process of the T2I model, we propose a swapping operation to mix well the prompt embeddings of a prompt pair for a novel object image generation, as shown in the left part of Figure \ref{fig:framework}. The swapping process is formalized as the following three steps:
\begin{itemize}
\item \textbf{Encoding} a prompt pair $(p_1,p_2)$ by using a text encoder, 
\begin{align}
E_1=\mathcal{E}(p_1)\in \mathbb{R}^{h\times w}  \ \ \  \text{and} \ \ \ E_2=\mathcal{E}(p_2)\in \mathbb{R}^{h\times w},
\label{eq:encoding}
\end{align}
\item \textbf{Swapping} intrinsic elements by using an exchanging vector $f \in \{0,1\}^{w}$, 
\begin{align}
E_f=E_1\text{diag}(f)+E_2\text{diag}(1-f) \in \mathbb{R}^{h\times w} ,
\label{eq:creativeprocess}
\end{align}
\item \textbf{Generating} a novel image by using an image generator, 
\begin{align}
I_f=\mathcal{G}(E_f) \in \mathbb{R}^{H\times W},
\label{eq:generating}
\end{align}
\end{itemize}      
where $\text{diag}(\cdot)$ is an operation to diagonalize a vector, and $f\in \{0,1\}^{w}$ is a binary vector to swap the column vectors of the prompt embeddings $E_1$ and $E_2$. (Neural swapping process, please see the \textit{Supp. Mat. \ref{sec:learningswapping}}.)

\begin{figure}[t]
\centering
\includegraphics[width=0.92\linewidth]{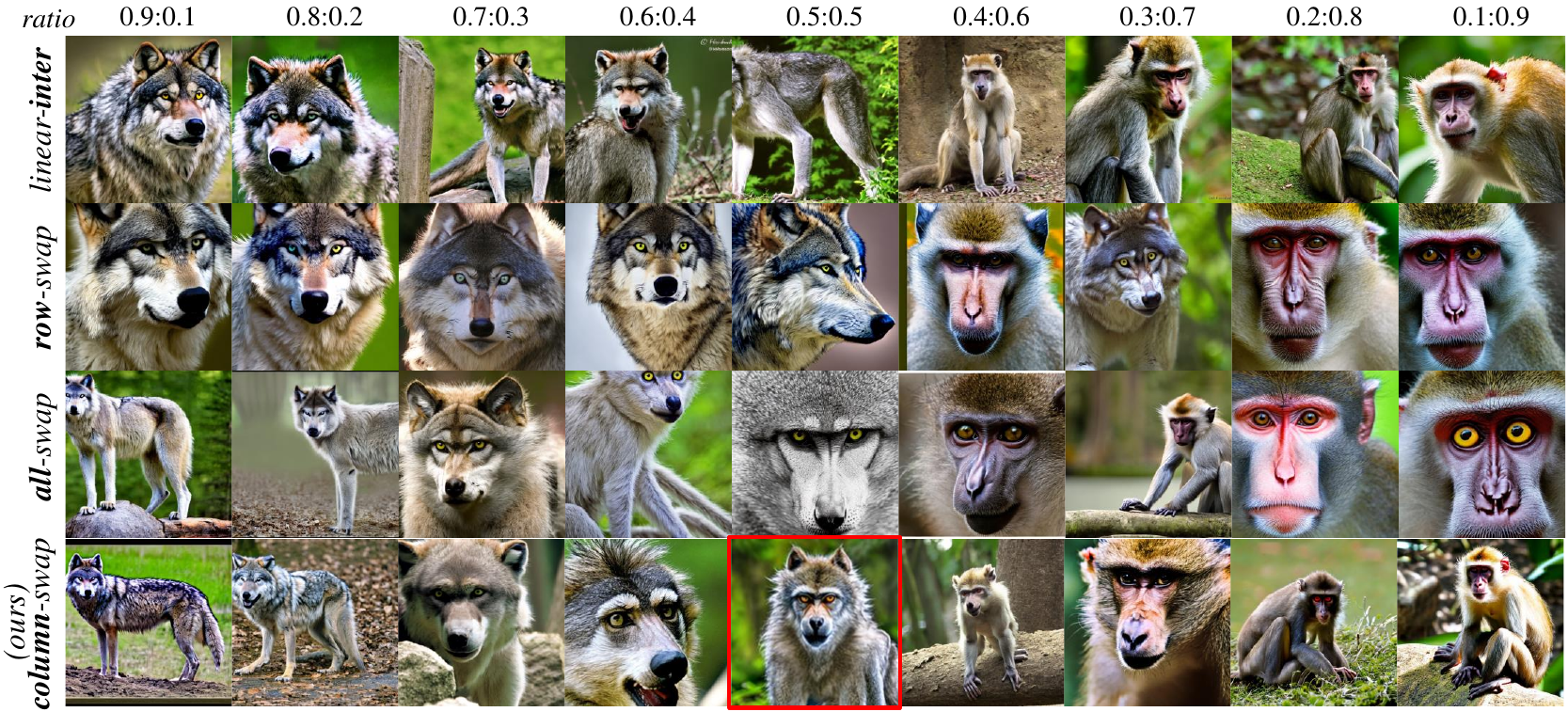}
\caption{Visualization comparison of different swapping ways.} 
\label{fig:cmp_swap}
\end{figure}
\textbf{Discussions on swapping ways between $E_1$ and $E_2$.} Here, we discuss four swapping ways to ingeniously merge the meaningful characteristics of these prompt embeddings, as shown in Figure \ref{fig:cmp_swap}. The first way is a simple and popular \textit{linear interpolation}, spanning a linear embedding space using $\alpha E_1 + (1-\alpha)E_2$, where $\alpha$ ranges from 0 to 1. Nevertheless, since this method mixes the whole embedding manifolds, it may struggle to blend the inherent characteristics of the embeddings $E_1$ and $E_2$ effectively, often resulting in predictable and unremarkable outcomes. The second way \textit{swaps the row vectors} of these embeddings, corresponding to words (objects). Similarly, it faces challenges in effectively exchanging intrinsic elements. To achieve a more surprising combination, the third way \textit{swaps all the corresponding elements} of $E_1$ and $E_2$. However, the computational cost of finding the swapping matrix makes it impractical. In contrast, the fourth way \textit{swaps the column vectors} of these embeddings, effectively mixing the inherent characteristics of the words. This increases the likelihood of creative combinations. Several experiments have demonstrated meaningful results by swapping the column vectors multiple times, as shown in Figures \ref{fig:creativeobject} and \ref{fig:result_compare}.

\subsection{Balance Region} 
Using the above swapping technique in the Equations \eqref{eq:encoding}-\eqref{eq:generating}, a new image $I_f$ is created by combining the text prompts $p_1$ and $p_2$. In this subsection, we establish a balance region to sample potentially high-quality combinations depending on only the images ($I_1$ and $I_2$) generated using $p_1$ and $p_2$ as anchor points, excluding human priors. This region adheres to a fundamental sense:
\begin{center}
\textit{A new image $I_f$ can be considered of potentially high-quality combination if it maintains an appropriate balance in distance from the anchor images $I_1$ and $I_2$.} 
\end{center}
This distance must adhere to three key rules. The first rule entails maintaining a balance between the distances from $I_f$ to $I_1$ and from $I_f$ to $I_2$, demonstrating an equilibrium between them. The second rule highlights that a substantial separation between these distances signals a higher likelihood of generating content that is disordered, devoid of meaningful. The third rule underscores that a minimal distance indicates that the generated image closely resembles the input data, potentially lacking novelty and surprise. In mathematical terms, we can formalize them into two distance criteria:
\begin{itemize}
\item \textbf{Balancing Distances:} Achieving a balance between the distances $d(I_f, I_1)$ and $d(I_f, I_2)$ through an inequality involving a constant $\alpha$, 
\begin{align}
\left|d(I_f,I_1)-d(I_f,I_2)\right|\leq\alpha,
\label{eq:balance}
\end{align}
\item \textbf{Controlling Bounds:} Constraining the upper bound of the average distance between $d(I_f,I_1)$ and $d(I_f,I_2)$ by using an inequality with a constant $\beta$, 
\begin{align}
d(I_f,I_1)+d(I_f,I_2)\leq 2\beta,
\label{eq:control}
\end{align}
\end{itemize}  
where $\alpha\geq0$, $\beta\geq d(I_1,I_2)/2\geq0$, $d(I_1,I_2)$ is the distance between the anchor images $I_1$ and $I_2$, $|\cdot|$ is the absolute value function, and $d(a,b)\geq 0$ is a positive function to compute the distance between $a$ and $b$. These two criteria govern a region in within which the combinatorial image $I_f$ has the potential to be potential high-quality. Now, let's delve into a geometrical analysis.

\begin{wrapfigure}{r}{0.47\textwidth}
\vskip -0.23in
\centering
\includegraphics[width=0.95\linewidth]{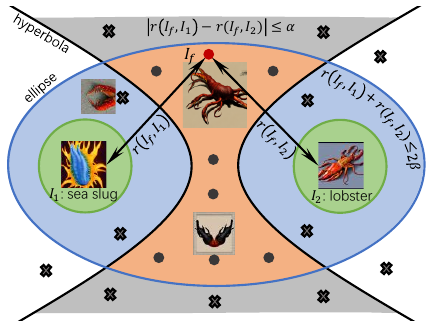}
\caption{Geometrical visualization of the high-quality composite image's potential orange region by balancing the distances between $I_f$ and the anchor images $I_1$, $I_2$.}
\label{fig:fig_creativeimage}
\vskip -0.25in
\end{wrapfigure}%
\textbf{Geometrical Explanations.} In defining the potential combinatorial space for image $I_f$, we employ geometric shapes—a hyperbola and an ellipse—as shown in Figure \ref{fig:fig_creativeimage}. The hyperbola constrains combinatorial images per Equation \eqref{eq:balance}: $\left|d(I_f,I_1)-d(I_f,I_2)\right|=\alpha$, where $\alpha$ is constant and $I_1$ and $I_2$ are fixed points, delineating the grey-shaded area. At $\alpha=0$, it forms a balanced perpendicular line between $I_1$ and $I_2$. Similarly, Equation \eqref{eq:control} defines an elliptical upper boundary: $\left(d(I_f,I_1)+d(I_f,I_2)\right)=2\beta$, with $I_1$ and $I_2$ fixed and $\beta$ constant, representing blue-shaded areas. At $\beta=d(I_1,I_2)/2$, it creates a line between $I_1$ and $I_2$. Considering both shapes simultaneously, $I_f$ falls within the overlapping orange region, indicating potential for high-quality blended images. Acceptable and rejected images are represented by circle and fork points, respectively.

While the balance region is not flawless, it offers control over $I_f$ quality to a degree in Figure \ref{fig:fig_creativeimage}. Additionally, green areas depict distributions of original images $I_1$ and $I_2$, contrasting with the orange region. This diagram shows that images within our balance region deviate from the data distribution, as evidenced by comparison with dataset retrievals in Figure \ref{fig:result_compare}.

\subsection{Balance Swap-Sampling} 

By combining the swapping technique with the balance region, we propose a balance swap-sampling method to sample a promising blend image $I_f$ on the prompt pair $(p_1,p_2)$. We first generate a set of $N$ randomly swapping vectors $\mathcal{F}$, and correspondingly produce an image set $\mathcal{I}$ using the Equations \eqref{eq:encoding}-\eqref{eq:generating}. Note that $f\in\mathcal{F}$ corresponds to $I_f\in\mathcal{I}$ one by one. Depending on the prompts $p_1,p_2$, and their generated images $I_1,I_2$, we introduce a coarse-to-fine sampling process comprising the following three steps.

\textbf{Coarse Sampling Using Semantic Distance.}
We select a coarse subset $\mathcal{I}_{\text{coarse}}$ from the set $\mathcal{I}$ by using a semantic balance, similar to the Equation \eqref{eq:balance}. This helps us maintain a balanced semantic content between the image $I_f$ and the text prompts $p_1$ and $p_2$. The coarse sampling is defined as:
\begin{align}
\mathcal{I}_{\text{coarse}} = \left\{I_f \ \boldsymbol{|} \left| d(I_f,p_1) - d(I_f,p_2) \right| \leq \theta, \ \ I_f\in\mathcal{I} \right\},
\label{eq:coarsesampling}
\end{align}
where $\theta$ is a width threshold of the semantic balance area. Here, we fix $\theta$ at 0.05.

\textbf{Fine Sampling Using Image Distances.} We further choose a fine subset $\mathcal{I}_{\text{fine}}$ from $\mathcal{I}_{\text{coarse}}$ by using the balance region in the Equations \eqref{eq:balance} and \eqref{eq:control}, ensuring a balanced relationship among $I_f$, $I_1$ and $I_2$. This leads to potential high-quality combinations. The fine sampling is expressed as:
\begin{align}
\mathcal{I}_{\text{fine}} =& \left\{I_f \ \boldsymbol{|}\left| d(I_f,I_1)-d(I_f,I_2)\right|\leq\alpha \ \& \right. \nonumber \\ 
& \left.  d(I_f,I_1)+d(I_f,I_2) \leq 2\beta, \ \ I_f\in\mathcal{I}_{\text{coarse}}  \right\},
\label{eq:finesampling}
\end{align}
where $\alpha=d_{\left\lceil |\mathcal{I}_{\text{coarse}}|\cdot\overline\alpha \right\rceil}^d$, $ \beta=d_{\lceil |\mathcal{I}_{\text{coarse}}|\cdot\overline\beta \rceil}^s$, $0\leq\overline\alpha\leq1$, $0\leq\overline\beta\leq1$, $\lceil\cdot\rceil$ denotes rounding up to the nearest integer, $|\mathcal{I}_{\text{coarse}}|$ represents the cardinality of the set $\mathcal{I}_{\text{coarse}}$, and $d^d_i$ and $d^s_i$ refer to the $i$-th element of the descendingly sorted sets $\mathcal{D}^d=\left\{d^d=\left| d(I_f, I_1) - d(I_f, I_2) \right|, I_f \in \mathcal{I}_{\text{coarse}} \right\}$ and $\mathcal{D}^s=\left\{d^s=d(I_f, I_1) + d(I_f, I_2),\right.$ $\left. I_f \in \mathcal{I}_{\text{coarse}} \right\}$, respectively. In this paper, we set $\overline\alpha=0.4$, and $\overline\beta=0.1$. The subset $\mathcal{I}_{\text{fine}}$ indicates images that may exhibit excellent mixing characteristics, denoted as black points within the orange region in Figure \ref{fig:fig_creativeimage}.

\textbf{Choosing the Optimal Image Using Segmentation Methods.} Unfortunately, $\mathcal{I}_{\text{fine}}$ is not directly employed for the ideal selection. Instead, we adopt the Segment Anything Model (SAM) \cite{Kirillov2023segany} to enhance the visual semantic components, thereby facilitating the selection of the optimal combinatorial image. The final selection is made by maximizing the following objective:
\begin{align} 
I_{\text{opt}}^{(p_1,p_2)} = &\underset{I_f\in\mathcal{I}_{\text{fine}}}{\arg \max } \left\{r(I_f,I_{1},I_{2})\right\}, \ \text{with} \ \label{eq:I_best} \\
&r(I_f,I_{1},I_{2})= \left(s(I_f,I_{1})+s(I_f,I_{2})\right)/2, \nonumber 
\end{align}
where $s(I_f,I_{i})=\frac{1}{|\mathcal{C}_i|\times|\mathcal{C}_f|}\underset{c_f\in \mathcal{C}_f,c_i\in \mathcal{C}_i}{\sum}d(c_f,c_i)$, $\mathcal{C}_i=\text{SAM}(I_i) (i=1,2)$, $\mathcal{C}_f=\text{SAM}(I_f)$, and $\text{SAM}(I)$ represents a collection of segmented components extracted from the image $I$ using SAM \cite{Kirillov2023segany}. Using Equation \eqref{eq:I_best}, we identify the optimal image from the $\mathcal{I}_{\text{fine}}$, as depicted by red points in Figure \ref{fig:fig_creativeimage}. This process culminates in the creation of a combinatorial  object image that is both promising and amazing. Overall, our BASS is outlined in \textbf{Algorithm \ref{alg:ass}}. Note that we define the distance function as $d(a, b) = \text{cos}(\phi(a), \phi(b))$, where $\text{cos}(\cdot, \cdot)$ represents a cosine similarity, and $\phi(\cdot)$ corresponds to the features using the CLIP model \cite{radford2021learning}. 

\begin{algorithm}[t]
\caption{Balance Swap-Sampling (BASS).}
\begin{algorithmic}[1]
\State \textbf{input:} prompt pair $(p_1,p_2)$, their images $I_1$, $I_2$;
\State \textbf{initialize:} $\theta=0.05$, $\overline\alpha=0.4$, $\overline\beta=0.1$, $N=200$;
\State Generate a set $\mathcal{F}$ of $N$ randomly swapping vectors;
\State Produce an image set $\mathcal{I}$ using Eqs. \eqref{eq:encoding}-\eqref{eq:generating} and $f\in\mathcal{F}$;
\State Sample a coarse subset $\mathcal{I}_{\text{coarse}}$ from $\mathcal{I}$ by Eq. \eqref{eq:coarsesampling} with $\theta$;
\State Sample a fine subset $\mathcal{I}_{\text{fine}}$ from $\mathcal{I}_{\text{coarse}}$ by Eq. \eqref{eq:finesampling} with $\overline\alpha$, $\overline\beta$;
\State  Choose the optimal image $I_{\text{opt}}^{(p_1,p_2)}$ from $\mathcal{I}_{\text{fine}}$ by Eq. \eqref{eq:I_best};
\State \textbf{output:} $I_{\text{opt}}^{(p_1,p_2)}$.
\end{algorithmic}
\label{alg:ass}
\end{algorithm}

\section{Experiments}
\subsection{Experimental Settings}

\textbf{Dataset.} To showcase the power of combinational creativity in T2I synthesis, we have curated a novel dataset comprising prompt pairs. We leveraged the vast vocabulary of ImageNet \cite{ILSVRC15}, consisting of 1,000 categories, to form our text set. From this collection, we randomly selected two distinct words to construct each prompt pair, such as \textit{macaque}-\textit{timber wolf}. Our dataset encompasses a total of 5075 prompt pairs, representing a fraction of the possible combinations. In addition, to compare the human artworks, we have gathered a selection of pieces created by \textit{Les Créatonautes} from the \href{https://www.instagram.com/les.creatonautes/}{Instagram}.

\textbf{Sampling Settings.} For the sampling process, the text encoder $\mathcal{E}(\cdot)$ and the image generator $\mathcal{G}(\cdot)$ were pretrained using the CLIP model with ViT-L/14@336p backbone \cite{radford2021learning} and Stable-Diffusion2 \cite{Rombach2022latentDM} or 
\href{https://clipdrop.co/stable-diffusion-turbo}{Stable-Diffusion XL turbo}, respectively. This CLIP model \cite{radford2021learning} was also employed in the distance function. The Segment Anything Model (SAM) \cite{Kirillov2023segany} served as the pre-trained segmentation method. We conducted our experiments using four NVIDIA GeForce RTX 3090 GPUs, with a batch size of 64 per GPU.

\textbf{Evaluation metrics.} Following the creative criteria, we evaluate the combinatorial object images generated by our method as follow. Firstly, we denote the cosine similarity between $a$ and $b$ using the CLIP model as $cos(a,b)$. From the generated image $I_f$ and the prompt pairs $(p_1,p_2)$ with their corresponding images $(I_1,I_2)$, we define text- and image-balance metrics as $|cos(I_f,p_1)-cos(I_f,p_2)|$ and $|cos(I_f,I_1)-cos(I_f,I_2)|$, respectively, to quantify the degree of balance. Smaller metrics indicate better balance. Additionally, we define text- and image-average similarities as $(cos(I_f,p_1)+cos(I_f,p_2))/2$ and $(cos(I_f,I_1)-cos(I_f,I_2))/2$, respectively, to measure novelty. Smaller similarities indicate higher novelty. Secondly, we randomly select two distinct categories from ImageNet \cite{ILSVRC15} to create \textit{surprising} prompt pairs.
Thirdly, we employ PickScore \cite{Kirstain2023PickaPicAO} and HPS-v2 \cite{wu2023hpsv2}, which fine-tuned their CLIP models using human-preference datasets, to evaluate their \textit{human-preference values}. Finally, we also conduct a user study to assess the creativity of our combinatorial objects.

\begin{figure}[t]
\centering
\includegraphics[width=0.97\linewidth]{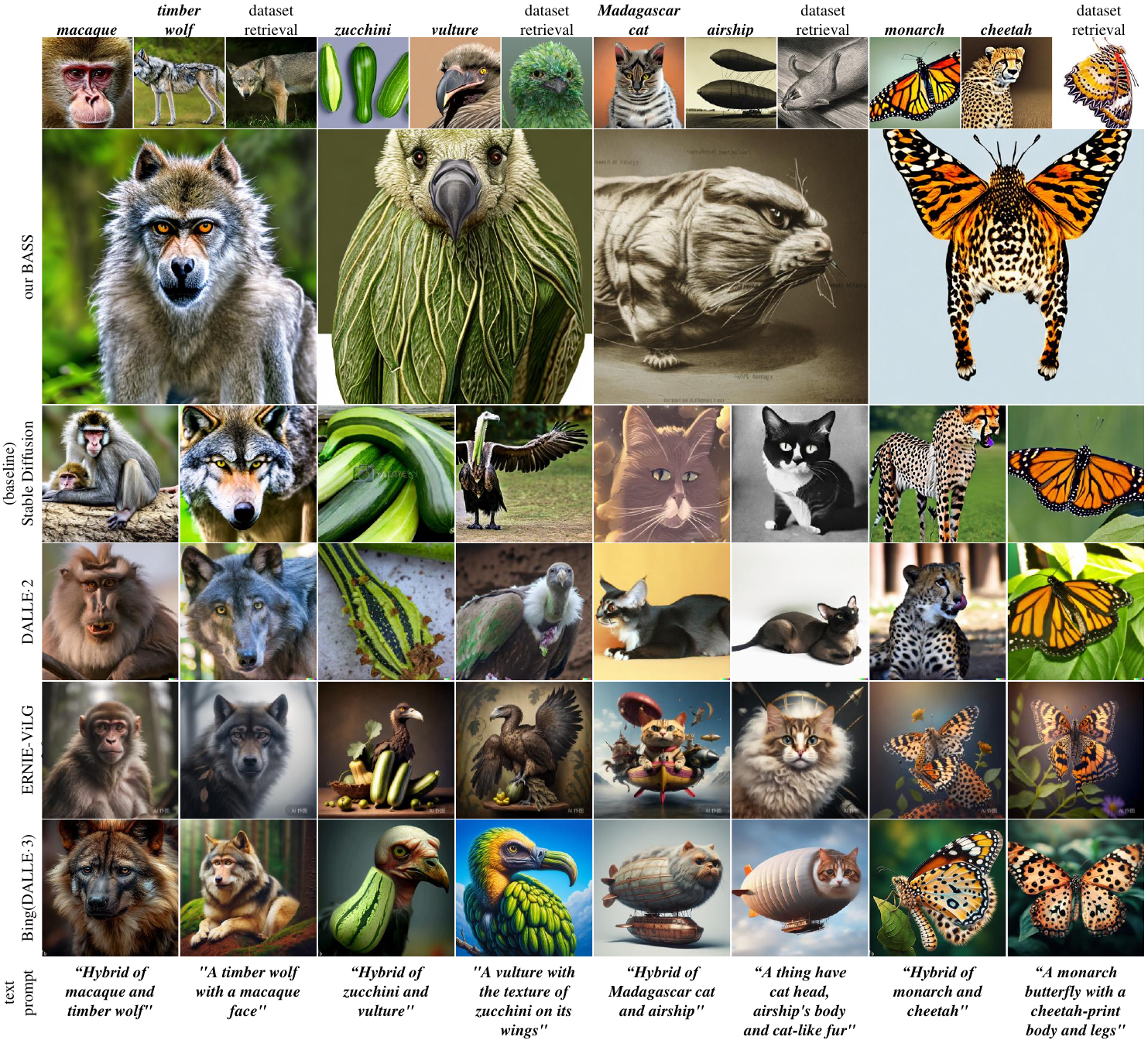}
\caption{Visual comparisons of combinatorial object generations. We compare our BASS with Stable-Diffusion2 \cite{Rombach2022latentDM}, DALLE2 \cite{Ramesh2022DALLE2}, ERNIE-ViLG2 (Baidu) \cite{Feng2022ERNIE-ViLG2} and Bing (Microsoft) using a hybrid prompt. 
For a fairer comparison, we incorporate detailed textual descriptions alongside our generated images as input prompts. However, these models have not achieved results closely aligned with our own, highlighting the superior creative combinatorial capabilities of our BASS. Furthermore, our results notably differ from images retrieved from the LAION-5B dataset \cite{schuhmann2022laion} in the first row, highlighting our model's capacity to produce out-of-distribution images.} 
\label{fig:result_compare}
\vskip -0.3in
\end{figure}

\subsection{Main Results}
We conducted a comprehensive comparison of our BASS method with some human artworks created by \textit{Les Créatonautes}, four prominent Text-to-Image (T2I) models (\ie, Stable-Diffusion2 \cite{Rombach2022latentDM}, DALLE2 \cite{Ramesh2022DALLE2}, ERNIE-ViLG2 \cite{Feng2022ERNIE-ViLG2} and Bing), and one  combinational method, (\ie, Magicmix \cite{Liew2022Magicmix}, which bears the closest relevance to our work). Note that as Magicmix currently does not offer an available software, we utilized its unofficial implementation \cite{magicmixcode}.

\textbf{Comparison with Artworks Created by Human Artists.} 
Figure \ref{fig:creativeobject} presents both human artworks (top row) and our creations (middle row) across corresponding category pairs. Surprisingly, our BASS exhibits the remarkable ability to generate novel and astonishing species unprecedented in reality, contrasting with the seemingly natural assembly in human artworks. For example, the fusion of an owl with the distinctive features of a tiger in our creation \textit{owl}-\textit{tiger} showcases a blend of owl attributes with the texture, feet, and eye of a tiger, while the human artwork merely substitutes the tiger's head for that of an owl. Unlike the human depiction of \textit{giraffe}-\textit{snail}, which maintains the snail's shell and substitutes its body with that of a giraffe, our result uniquely features a snail adorned with the giraffe's head and legs, as depicted in Figure \ref{fig:creativeobject}. Furthermore, we enlisted the opinions of 116 non-artists to evaluate eight artworks alongside our creations. From this study, we amassed a total of 928 votes, with 71.3\% of non-artists expressing a preference for our creations. For additional comparisons and detailed information on the non-artists study, please refer to \textit{Supp. Mat. \ref{sec:comparsionwithartworks}}.

\textbf{Comparison with the SOTA T2I models.} 
Figure \ref{fig:result_compare} showcase several examples of image generation achieved through those models using prompt pairs. We make the following three observations.
First, compared to the SOTA T2I models, our BASS exhibits a stronger ability to generate a creative object by inputting two different objects. Although the images generated by other models are colorful and rich in detail, they do not fully display the mixed features of the two objects, as seen in \textit{macaque}-\textit{timber wolf} and \textit{zucchini}-\textit{vulture} in Figure \ref{fig:result_compare}.
Second, to evaluate the Out-Of-Distribution (OOD) generation ability of our BASS, we conducted a retrieval on the entire LAION-5B dataset \cite{schuhmann2022laion} to find the most similar image in the first row of Figure \ref{fig:result_compare}. By comparing our created images with the retrieved images, we found that they significantly differ from the retrieved ones, highlighting the distinctiveness of our BASS's output. Note that our BASS generation differs from outlier \cite{Du2023outlier} and guidance \cite{Zhang2023gif} imaginations 
as they generate the data in the same category. Third, when comparing the images generated using intricate textual descriptions in conjunction with our created images as input prompts, these models still struggles to create plausible compositions, such as \textit{A timber wolf with a macaque face} in the Figure \ref{fig:result_compare}. While Bing employs \textit{A vulture with the texture of zucchini on its wings} to generate an object that looks beautiful, it utilizes entire zucchinis as the vulture's wings, rather than utilizing their texture. More results, refer to \textit{Supp. Mat. \ref{sec:moreresults}}.

\begin{wrapfigure}{r}{0.43\textwidth}
\vskip -0.0in
\centering
\includegraphics[width=0.97\linewidth]{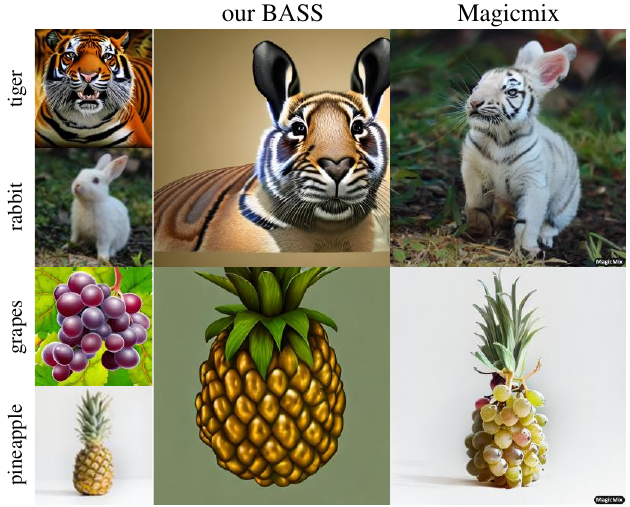}
\vskip -0in
\caption{Combinational comparison.}
\label{fig:magicmix_main}
\vskip -0.3in
\end{wrapfigure}

\textbf{Comparison with Magicmix \cite{Liew2022Magicmix}.} As it works with image-text inputs, our BASS utilizes both the image category and accompanying text to create a text pair. Figure \ref{fig:magicmix_main} shows the resulting combinational object images, observing that 
our BASS excels in seamlessly blending the features of both tiger and rabbit, incorporating elements like the rabbit's ears and the tiger's face more naturally. In contrast, Magicmix merely overlays the tiger's texture onto the rabbit, resulting in an unnatural appearance. More results, see \textit{Supp. Mat. \ref{sec:comparsionwithmagicmix}}.

\textbf{Evaluating novelty, surprise and value.}
Since we randomly choose two distinct categories from ImageNet \cite{ILSVRC15} to form text pairs, creating seemingly impossible associations, this selection directly confirms the \textit{associated surprise} of potential combinations, such as \textit{zucchini-vulture}, \textit{frog-broccoli}, and \textit{monarch-cheetah}. Our focus here is on assessing novelty and value.

\begin{wraptable}{r}{0.57\textwidth}
\vskip -0.45in
\centering
\setlength{\tabcolsep}{7pt}
\renewcommand{\arraystretch}{1.1}
\caption{Quantitative comparisons.}
\vskip -0in
\resizebox{0.97\linewidth}{!}{
\begin{tabular}{c|c|c|c|c} 
\Xhline{1.2pt} 
\multirow{2}{*}{\diagbox{Model}{Score}} & \multicolumn{2}{c|}{text-} & \multicolumn{2}{c}{image-} \\ 
\cline{2-5}
   & avg. sim.$ \downarrow $   & balance$ \downarrow $ & avg. sim.$ \downarrow $   & balance$ \downarrow $        \\ 
\hline
our BASS  & 0.328 & 0.061 & 0.543 & 0.126     \\
MagicMix \cite{Liew2022Magicmix} & 0.282 & 0.100  & 0.536 & 0.220  \\
\Xhline{1.2pt} 
\end{tabular}}
\label{tab:cmp_Magicmix}
\vskip -0.25in
\end{wraptable}
\textit{Balanced novelty.} Table \ref{tab:cmp_Magicmix} reports the metrics for text- and image-balance, as well as text- and image-average similarities. Our observations are as follows: 1) Our BASS exhibits superior balance compared to MagicMix \cite{Liew2022Magicmix}. 2) Both BASS and MagicMix demonstrate lower similarity, evidenced by values lower than the similarity between the text prompt and its generated image, which consistently approaches 1. 3) While MagicMix slightly outperforms our BASS in novelty, its visual outputs, in Figure \ref{fig:magicmix_main}, are less intuitive than ours.

\begin{wrapfigure}{r}{0.47\textwidth}
\vskip -0.27in
\centering
\includegraphics[width=0.97\linewidth]{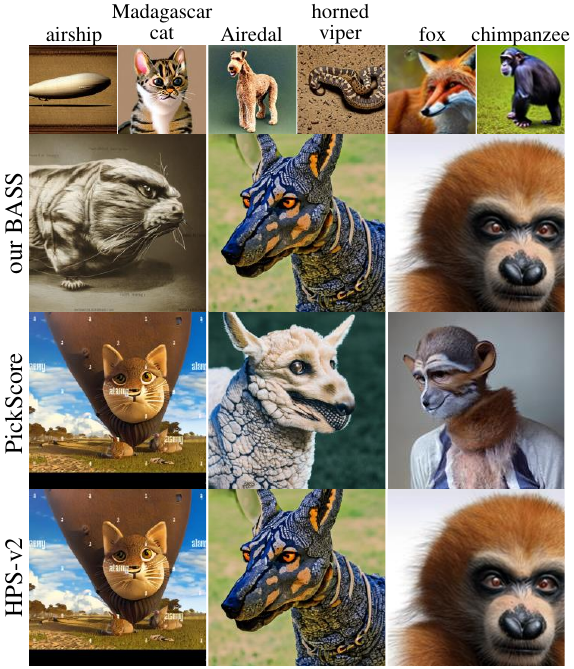}
\caption{Sampling visualizations compared our BASS with HPSv2 and PickScore. }
\label{fig:result_HPT}
\vskip -0.25in
\end{wrapfigure}%
\textit{Human-preference value.} To assess the human-like superiority of our BASS method, we employed PickScore \cite{Kirstain2023PickaPicAO} and HPSv2 \cite{wu2023hpsv2} to calculate evaluation scores for selecting the optimal image with the highest score from the new image set $\mathcal{I}$. Subsequently, we utilized these metrics to evaluate the optimal images, and the findings are presented in Table \ref{tab:Quantitativeevaluations}. Our results indicate that our BASS method closely rivals the performance of PickScore and HPSv2. Notably, our method operates independently of human intervention, except for the selection of hyper-parameters using HPSv2. In contrast, both PickScore and HPSv2 rely on human-preference datasets to fine-tune the CLIP models for T2I model evaluation. This demonstrates our BASS possesses a capability with \textit{comparable human-preference value}. Furthermore, the sampling examples are illustrated in Figure \ref{fig:result_HPT}. Our BASS method excels in generating superior combinatorial images compared to both HPSv2 and PickScore.
\begin{minipage}{\textwidth}
\vskip 0.1in
\begin{minipage}[t]{.41\textwidth}
\centering
\setlength{\tabcolsep}{3pt}
\renewcommand{\arraystretch}{0.97}
\makeatletter\def\@captype{table}
\caption{Quantitative comparisons.}
\vskip -0.07in
\resizebox{0.97\linewidth}{!}{
\begin{tabular}{c||c|c|c}
\Xhline{1.2pt} 
Models & PickScore \cite{Kirstain2023PickaPicAO} & HPSv2 \cite{wu2023hpsv2} & Our BASS \\ 
\hline
PickScore $ \uparrow $  &  0.207 & 0.202 & 0.200  \\
HPSv2 $ \uparrow $  &  0.246 & 0.253 & 0.242  \\
\Xhline{1.2pt}
\end{tabular}}
\label{tab:Quantitativeevaluations}
\end{minipage}
\begin{minipage}[t]{.55\textwidth}
\centering
\setlength{\tabcolsep}{3pt}
\renewcommand{\arraystretch}{1.2}
\makeatletter\def\@captype{table}
\caption{User study of combinational creations.}
\vskip -0.07in
\resizebox{0.97\linewidth}{!}{
\begin{tabular}{c||c|c|c|c|c}
\Xhline{1.2pt} 
Models & Our BASS & \small{\makecell[c]{SD2 \cite{Rombach2022latentDM}  \\ (baseline)}}  & DALLE2 \cite{Ramesh2022DALLE2} & ERNIE-ViLG2 \cite{Feng2022ERNIE-ViLG2}  & Bing  \\ 
\hline
Vote $ \uparrow $  &  \textbf{658} & 98 & 59 & 49 & 196  \\
\Xhline{1.2pt}
\end{tabular}}
\label{tab:votes}
\end{minipage}
\vspace{ 7pt}
\end{minipage}

\textbf{User Study.} We conducted a user study to assess the combinational creativity of our model against four other T2I methods in Table \ref{tab:votes} and \textit{Supp. Mat. \ref{sec:UserStudy}}. Each user evaluated 10 prompt pairs, resulting in 1,060 votes from 106 users. Our model garnered the highest number of votes, with 62\% of users favoring its creative outputs. The Bing model attracted interest from 18.5\% of users, while DALLE2 \cite{Ramesh2022DALLE2} received only 5\% of the votes despite its structural similarities to the Bing model. Stable-Diffusion2 \cite{Rombach2022latentDM} received 9\% of the votes, while ERNIE-ViLG2 \cite{Feng2022ERNIE-ViLG2} was preferred by only 4.6\% of users.

\textbf{Computational Costs.} In contrast to the baseline (Stable-Diffusion2), our additional models include CLIP and SAM. For a single text pair, we parallelize the data to complete the inter-evaluation process in 10 minutes using 4 RTX 4090 GPUs. \textit{When using Stable-Diffusion XL turbo in the same setting, it only requires a mere \textbf{40 seconds} to discover meaningful object images}.

\begin{wraptable}[]{r}{2.7in}
\vskip -0.47in
\begin{minipage}{.52\textwidth}
\centering
\setlength{\tabcolsep}{7pt}
\renewcommand{\arraystretch}{1.1}
\caption{Parameter analysis with $\theta$ using average HPSv2 scores \cite{wu2023hpsv2} of 20 text pairs. $+\infty$ represents all sampling images.}
\vskip -0in
\resizebox{1\linewidth}{!}{
\begin{tabular}{c||c|c|c|c|c}
\Xhline{1.2pt}
$\theta$ & 0.01 & 0.02 & \textbf{0.05}& 0.1 & $+\infty$  \\
\hline
HPSv2 & 0.2444 & 0.2451 & \textbf{0.2458} & 0.2392 &  0.2361    \\
\Xhline{1.2pt}
\end{tabular}}
\label{tab:theta}
\end{minipage}
\begin{minipage}{.55\textwidth}
\vskip -0.1in
\centering
\setlength{\tabcolsep}{7pt}
\renewcommand{\arraystretch}{1.1}
\caption{Parameter analysis with $\overline\alpha$ and $\overline\beta$ using average HPSv2 scores \cite{wu2023hpsv2} of 20 text pairs. }
\vskip -0in
\resizebox{0.87\linewidth}{!}{
\begin{tabular}{c|c|c|c|c} 
    \Xhline{1.2pt}
    $\overline\alpha$$\backslash$$\overline\beta$ & 0 & 0.2 & \textbf{0.4} & 0.6 \\
    \hline
    0 & 0.241 & 0.231 & 0.230& 0.223  \\
    \textbf{0.1} & 0.231 &0.242 & \textbf{0.243} & 0.239\\
    0.2 & 0.242 & 0.230 &0.237 & 0.224  \\
    0.3 & 0.240 & 0.231 &0.235 & 0.225  \\
    \Xhline{1.2pt}
    \end{tabular}}
\label{tab:alphabeta}
\end{minipage}
\vskip -0.25in
\end{wraptable} 
\subsection{Parameter Analysis and Ablation Study.} 
\textbf{Parameter Analysis.} During the sampling process, we determined the parameters $\theta$ in the Equation \eqref{eq:coarsesampling}, and $\overline\alpha$, and $\overline\beta$ in the Equation \eqref{eq:finesampling} using 20 text pairs. To 
begin, for each prompt pair $(p_1,p_2)$, we produce an image set $\mathcal{I}$ by randomly generating a set $\mathcal{F}$ consisting of $N=200$ swapping vectors. From $\mathcal{I}$, we coarsely select a subset $\mathcal{I}_{\text{coarse}}$ using Equation \eqref{eq:coarsesampling}, and the parameter $\theta$ is set to 0.05 by choosing the best average HPSv2 score \cite{wu2023hpsv2} in Table \ref{tab:theta}. This reduces the size of $\mathcal{I}_{\text{coarse}}$ approximates to 150. Next, we finely choose a subset $\mathcal{I}_{\text{fine}}$ from $\mathcal{I}_{\text{coarse}}$ using the Equation \eqref{eq:finesampling}. 
We set that $\overline\alpha=0.4$ and $\overline\beta=0.1$ by selecting the best average HPSv2 score in Table \ref{tab:alphabeta}. This 
reduces the size of $\mathcal{I}_{\text{fine}}$ to around 10. Finally, we obtain the optimal image $I_{\text{opt}}^{(p_1,p_2)}$ by maximizing the problem in the Equation \eqref{eq:I_best}. For illustrations of the sampled images using different $\theta$, $\overline\alpha$, and $\overline\beta$, refer to \textit{Supp. Mat. \ref{sec:ParameterAnalysis}}.

\textbf{Ablation Study of SAM.} As SAM \cite{Kirillov2023segany} is used to enhance the greater similarity in semantic components in semantic components with the original objects $I_1$ and $I_2$ within the final sampling set $\mathcal{I}_{\text{fine}}$, we conducted an ablation study to show the effectiveness of SAM in Figure \ref{fig:ablationstudy}. Using SAM, we can choose better conbinational object images.

\subsection{Discussions} 
We mainly discuss the potential applications and limitation of our BASS.

\textbf{Potential applications.} Firstly, we excel at crafting innovative and captivating animated characters for the entertainment and film industry (\eg, \textit{monarch-cheetah} in Figure \ref{fig:result_compare}).
Secondly, we possess the capability to generate imaginative artworks suitable for art design (\eg, \textit{frog-broccoli} and \textit{turtle-cat} in Figures \ref{fig:creativeobject}). For example, \textit{Les Créatonautes} designs the visual association between completely different objects.
Thirdly, our method extends the scope of the \textit{Out Of Distribution} (OOD) task to a new generation (see Figure \ref{fig:result_compare}).
Lastly, we pave the way for translating qualitative descriptions of creativity into achievable technologies within the realm of computer vision.

\begin{figure}[t]
\vskip -0in
\begin{minipage}[t]{.41\textwidth}
\vskip -1.32in
\centering
\includegraphics[width=0.87\linewidth]{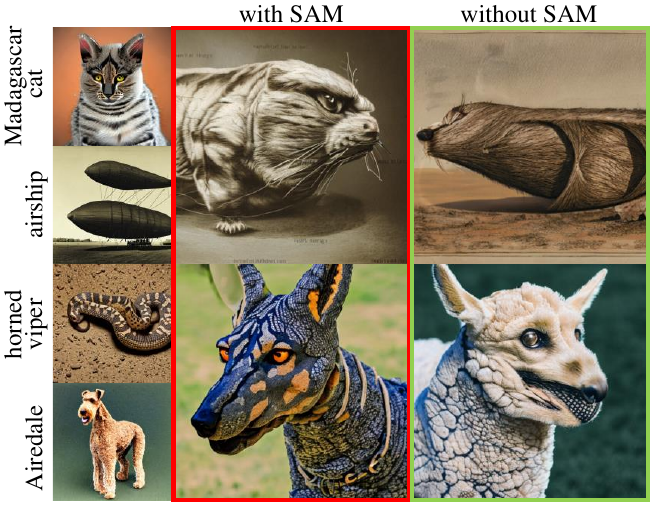}
\caption{Ablation study of SAM.}
\label{fig:ablationstudy}
\vskip -0in
\end{minipage}
\begin{minipage}[t]{.51\textwidth}
 \centering
\includegraphics[width=0.97\linewidth]{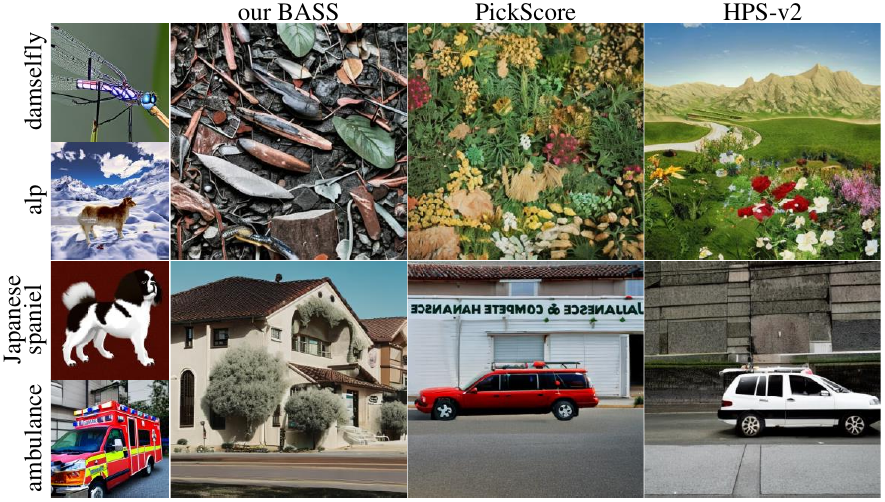}
\caption{Failure samples in sampling stage.}
\label{fig:failure}
\vskip -1.27in
\end{minipage}
\vskip -0.15in
\end{figure}

\textbf{Limitation.} Our method, despite its strengths, has a limitation: the balance swapping region can sometimes result in non-meaningful or chaotic images. The high-quality balance area in an unsupervised manner continues to pose a challenging problem that necessitates further investigation, as shown in Figure \ref{fig:failure}. We have included more failure examples in the \textit{Supp. Mat. \ref{sec:failures}} for reference. 

Note that for comparisons with more intricate prompts and generalizations involving three concepts, please refer to \textit{Supp. Mats. \ref{sec:intricateprompts} and \ref{sec:moreconcepts}}, respectively.

\section{Conclusion}

We have incorporated a simple sampling method into text-to-image synthesis, proposing a balance swap-sampling schema to generate meaningful objects by combining seemingly unrelated object concepts. Our first idea involves a swapping process that exchanges important information from two given prompts, creating fresh object images that go beyond these prompts and the original data distribution, enhancing novelty. Additionally, we introduce a balance region based on the CLIP metric, balancing the distance among the given prompts, original image generations, and our creations to sample high-quality combinatorial object images. We further employ the segment anything model to enhance the visual semantic components to select the optimal combinatorial image. Experimental results demonstrate that our approach surpasses popular T2I models in generating creative combinatorial objects, comparable to artworks by Les Créatonautes, a French creative agency.

\section*{Acknowledgements}
This work was partially supported by the National Science Fund of China, Grant Nos. 62072242 and 62361166670. We sincerely thank the French artist Les Creatonautes for granting us permission to use their images.

%
%
\bibliographystyle{splncs04}
\bibliography{egmain}

\clearpage
\appendix

\section{Learning a Neural Swapping Network}
\label{sec:learningswapping}
In this subsection, we learn a neural swapping network to generate the meaningful combinatorial object images with different styles. The swapping part in the subsection \ref{sec:swapping} is rewritten as follows:

\textbf{Swapping} their column vectors by learning a neural swapping vector, 
\begin{align}
E_f=E_1\text{diag}(f)+E_2\text{diag}(1-f),
\label{eq:creativeprocess}
\end{align}
where $f=S(\text{cat}(E_1,E_2);\psi)\in \{0,1\}^{w \times 1}$ is a neural swapping network from the concatenated embedding, $\text{cat}(E_1,E_2)$, to a binary output that consists of three convolutional layers and two fully connected layers with the parameter $\psi$. The architecture of the swapping network consists of three convolutional layers with a $3\times3$ kernel size, followed by two fully connected layers.

\textbf{Training Loss.} Using our BASS method in \textbf{Algorithm \ref{alg:ass}}, we can get the optimal combinatorial image $I_{\text{opt}}^{(p_1,p_2)}$, and then find its swapping vector $f_{\text{opt}}^{(p_1,p_2)}$. Based on the neural swapping network $f=S(\text{cat}(E_1,E_2);\psi)$ in Eq. \ref{eq:creativeprocess} and the optimal swapping vector $f_{\text{opt}}^{(p_1,p_2)}$, the training loss is defined as:
\begin{align}
L=\frac{1}{|\mathcal{P}|} \sum_{(p_1,p_2)\in\mathcal{P}} \|f_{\text{opt}}^{(p_1,p_2)}-S(\text{cat}(E_1,E_2);\psi)\|_2,
\label{eq:trainingloss}
\end{align}
where $\mathcal{P}$ is a set of the prompt pairs $(p_1,p_2)$ corresponding to the text pairs $(t_1,t_2)$, and $|\mathcal{P}|$ represents the cardinality of the set $\mathcal{P}$. To ensure training stability, we employed the RMSprop \cite{hinton2012neural} optimizer. 

\begin{figure}[h!]
    \centering
    \vspace{-15pt}
    \includegraphics[width=0.87\linewidth]{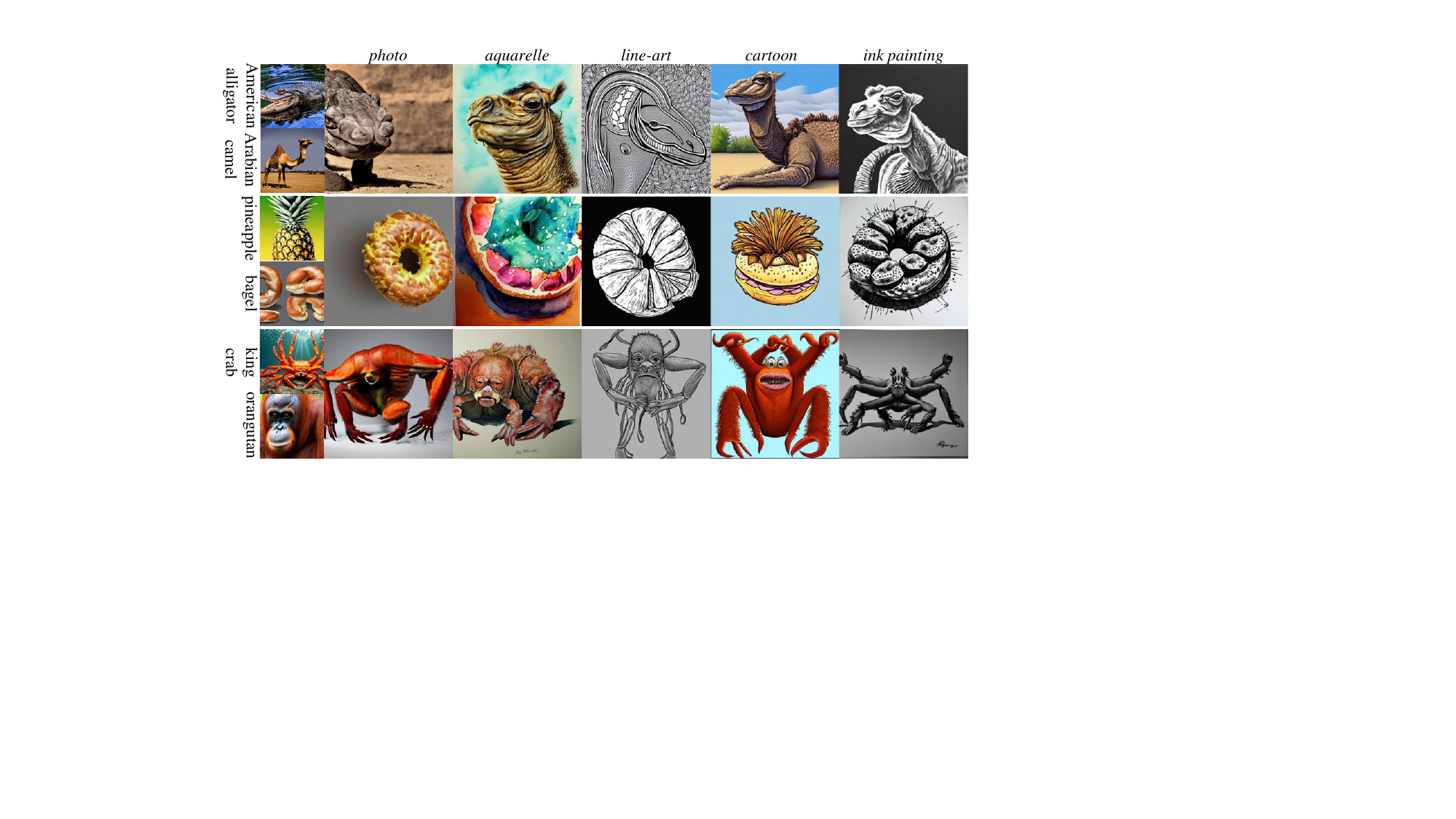}
    \vskip-0.1in
    \caption{Generalizations using four different styles including \textit{aquarelle}, \textit{line-art}, \textit{cartoon}, and \textit{ink painting}. It can observe that our results are still novel and surprising.}
    \label{fig:result_style}
    \vspace{-15pt}
\end{figure}

\textbf{Generalizations using different styles.} 
To evaluate the generalizations of our model, we expand its capabilities to four additional styles: \textit{aquarelle}, \textit{line-art}, \textit{cartoon}, and \textit{ink painting} for text-to-image generation. We achieve this without the need for excessive sampling and training. The results in Figure \ref{fig:result_style}, demonstrate that our model maintains its impressive creativity by generating novel species across these different styles, such as \textit{(orangutan, king
crab)}.

\section{Comparisons with Human Artworks}
\label{sec:comparsionwithartworks}

\begin{figure}
    \centering
    \vskip -0.3in
    \includegraphics[width=0.87\linewidth]{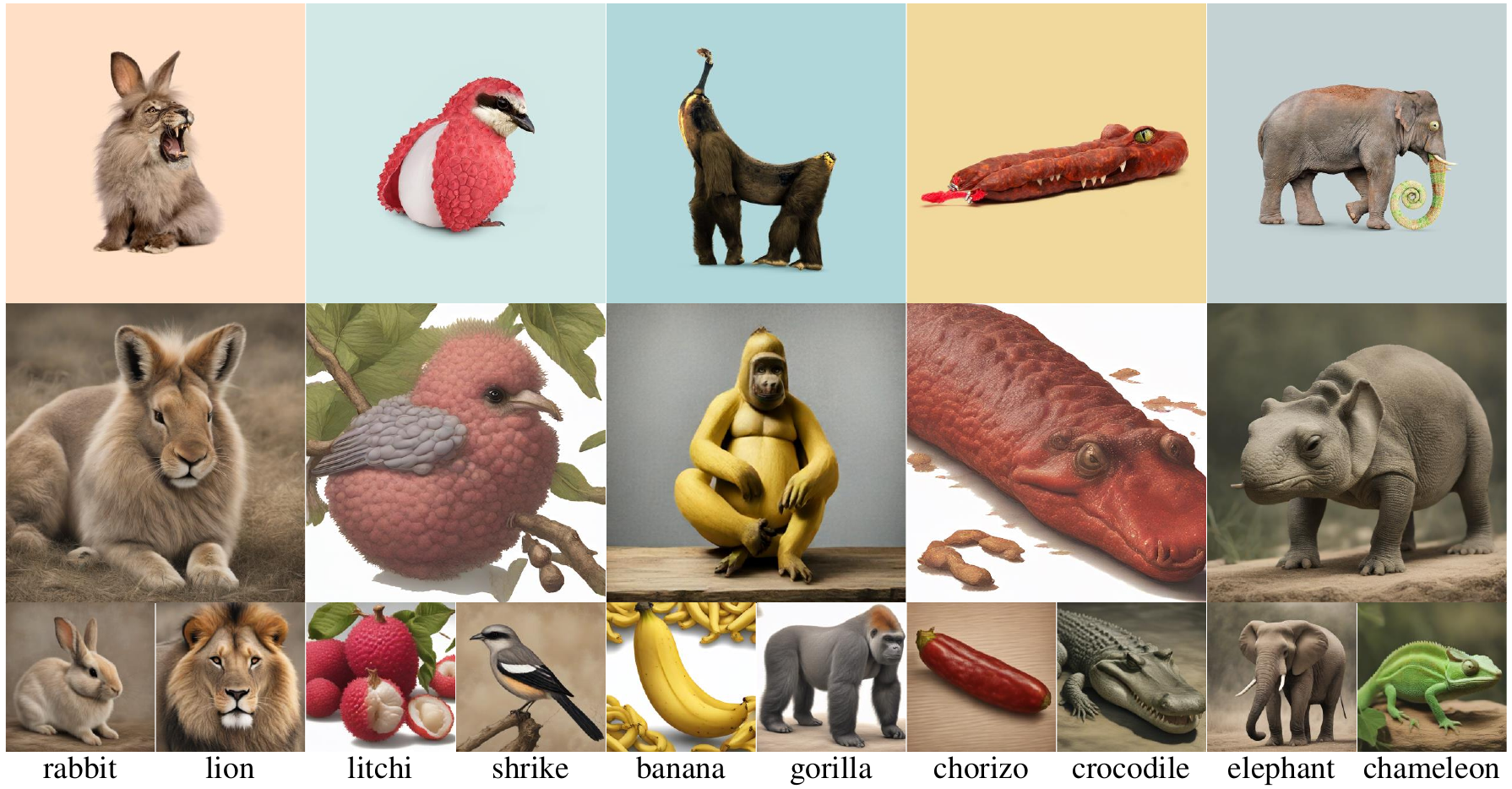}
    \vskip-0.1in
    \caption{More visualization results of artist comparison.}
    \label{fig:art_more}
    \vskip -0.25in
\end{figure}

\begin{wraptable}{r}{0.57\textwidth}
\centering
\vskip -0.45in
\setlength{\tabcolsep}{7pt}
\renewcommand{\arraystretch}{1.1}
\caption{Quantitative comparisons with artist.}
\resizebox{0.97\linewidth}{!}{
\begin{tabular}{c|c|c|c|c} 
\Xhline{1.2pt} 
\multirow{2}{*}{\diagbox{Model}{Score}} & \multicolumn{2}{c|}{text-} & \multicolumn{2}{c}{image-} \\ 
\cline{2-5}
   & avg. sim.$ \downarrow $   & balance$ \downarrow $ & avg. sim.$ \downarrow $   & balance$ \downarrow $        \\ 
\hline
our BASS  & 0.351 & 0.037 & 0.647 & 0.016     \\
artist    & 0.350 & 0.054 & 0.572 & 0.083  \\
\Xhline{1.2pt} 
\end{tabular}}
\label{tab:cmp_art}
\vskip -0.25in
\end{wraptable}

In this section, we provide more results of comparison with human artworks. In Figure \ref{fig:art_more}, we present additional comparison results between our method and images created by human artists. Similarly, the top row corresponds to images created by artists, the middle row to our results, and the bottom row to the prompts and corresponding image results used by both us and the artists. The results shown in the figure also illustrate that the fusion of our results tends to lean towards creating a new object that simultaneously incorporates features from both original prompts, rather than a simple fusion of two separate objects.
Similarly, we conducted quantitative balanced-novelty calculations, as shown in Table \ref{tab:cmp_art}. Our results outperform those of the artists in terms of balance.

\begin{figure}
    \centering
    \vskip -0in
    \includegraphics[width=0.87\linewidth]{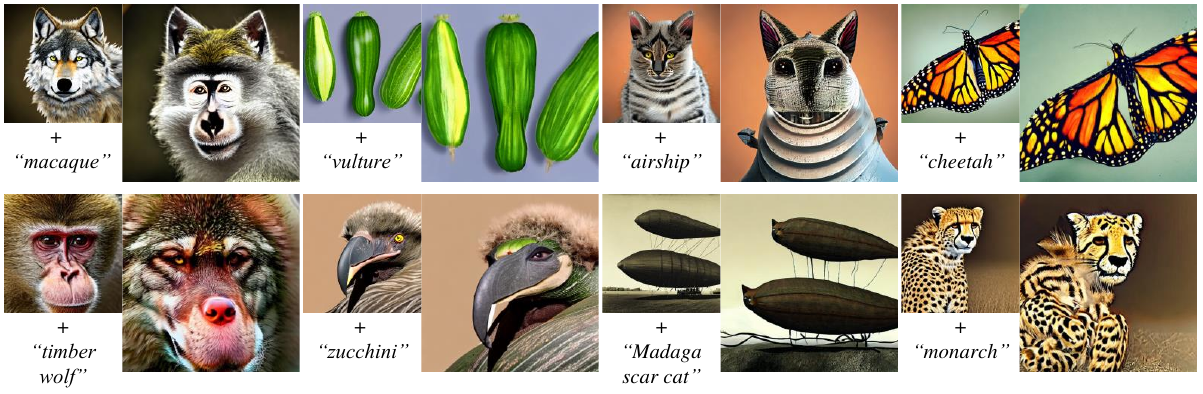}
    \vskip-0.1in
    \caption{Generalizations using unofficial code \cite{magicmixcode} of Magicmix with prompt-pairs in Figure \ref{fig:result_compare}.}
    \label{fig:magicmix_1}
    \vskip -0.2in
\end{figure}

\section{Comparisons with Magicmix}
\label{sec:comparsionwithmagicmix}

Here, we provide additional comparative results of our findings compared to those presented in Magicmix \cite{Liew2022Magicmix}, except for Figure \ref{fig:magicmix_main}. As there is no open-source software available for Magicmix, we employed an unofficial implementation \cite{magicmixcode}. Utilizing the four text pairs depicted in Figure \ref{fig:result_compare} to generate eight text-image pairs as inputs for Magicmix, the outcomes are illustrated in Figure \ref{fig:magicmix_1}. It is evident that our combined results significantly outperform those produced by Magicmix.

\begin{figure}[t]
    \centering
    \includegraphics[width=0.87\linewidth]{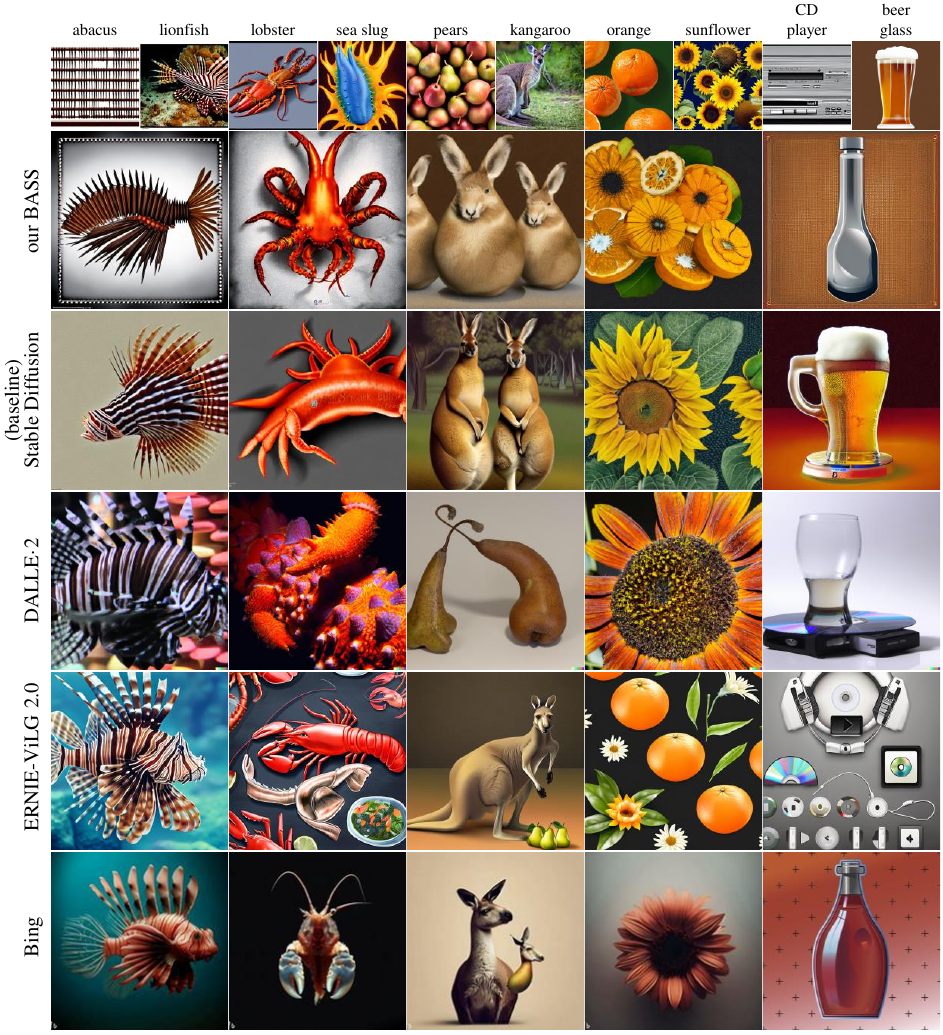}
    \caption{More visualization results of the User Study.}
    \label{fig:sup_usrstu}
    \vspace{-15pt}
\end{figure}

\begin{figure}[t]
    \centering
    \vskip -0in
    \includegraphics[width=0.7\linewidth]{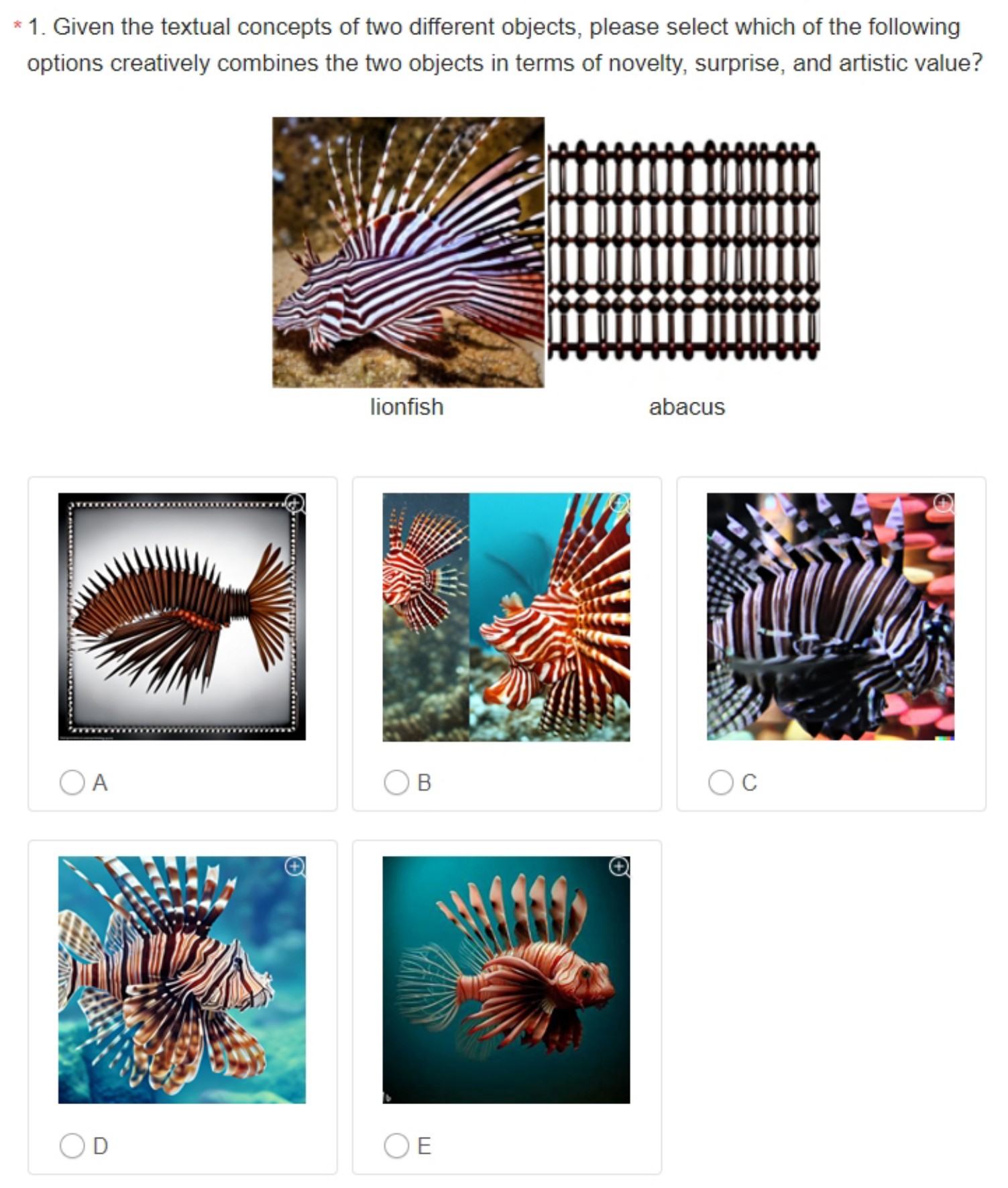}
    \vskip-0.1in
    \caption{Interface of the User Study.}
    \label{fig:user_study_inter}
    \vskip -0.15in
\end{figure}

\begin{table}[ht]
    \linespread{0.8}
    \centering
    \setlength{\tabcolsep}{6pt}
    \vskip -0in
    \renewcommand{\arraystretch}{1.1}
    \caption{The proportion of each option in each question in the User Study.}
    \resizebox{1\linewidth}{!}{
    \begin{tabular}{c||ccccc}
    \Xhline{1.2pt}
    \diagbox{prompt-pairs}{options(Models)}& A(our BASS) & B(SD2\cite{Rombach2022latentDM}-baseline) & C(DALLE·2\cite{Ramesh2022DALLE2}) & D(ERNIE-ViLG2\cite{Feng2022ERNIE-ViLG2}) & E(Bing) \\ 
    \hline
    lionfish and abacus          & 45.28\%    & 13.21\%     & 10.38\%    & 18.87\%        & 12.26\% \\
    lobster and sea slug         & 62.26\%    & 12.26\%     & 12.26\%    & 3.77\%         & 9.43\%  \\
    kangaroo and pears           & 64.15\%    & 20.75\%     & 5.66\%     & 4.72\%         & 4.72\%  \\
    sunflower and orange         & 75.47\%    & 15.09\%     & 1.89\%     & 2.83\%         & 4.72\%  \\
    macaque and timber wolf      & 75.47\%    & 2.83\%      & 2.83\%     & 3.77\%         & 15.09\% \\
    Australian terrier and tiger & 51.89\%    & 12.26\%     & 1.89\%     & 3.77\%         & 30.19\% \\
    toucan and bathing cap       & 27.36\%    & 3.77\%      & 4.72\%     & 2.83\%         & 61.32\% \\
    zucchini and vulture         & 85.85\%    & 0.94\%      & 1.89\%     & 1.89\%         & 9.43\%  \\
    jackfruit and thresher       & 79.25\%    & 0.94\%      & 3.77\%     & 1.89\%         & 14.15\% \\
    CD player and beer glass      & 53.77\%    & 10.38\%     & 10.38\%    & 1.89\%         & 23.58\% \\ 
    \Xhline{1.2pt}
    \end{tabular}
    }
    \label{tab:sup_usrstu}
    \vskip -0in
\end{table}

\begin{figure}[ht]
    \begin{minipage}[t]{0.45\linewidth}
        \centering
        \includegraphics[width=0.9\linewidth]{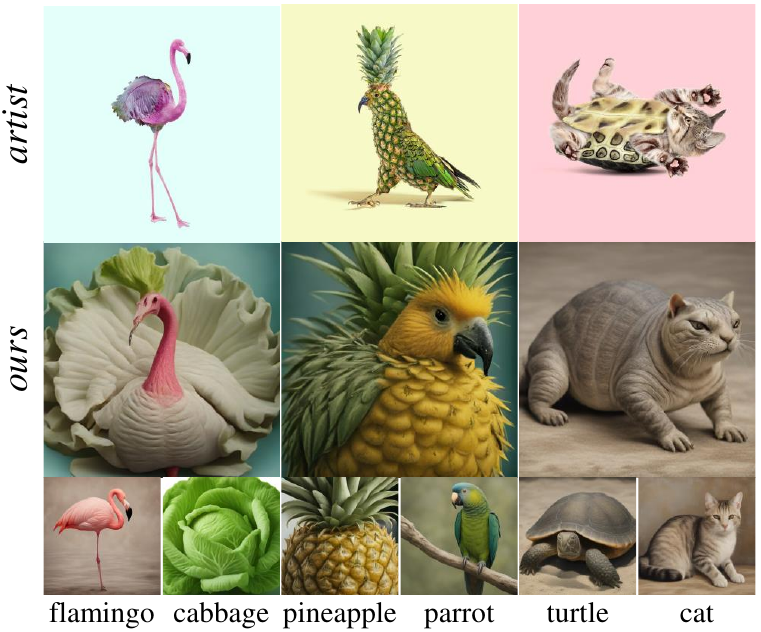}
        \caption{More visualization results of artist comparison User Study.}
        \label{fig:user_stu_art}
    \end{minipage}
    \begin{minipage}[t]{0.55\linewidth}
        \vspace{-1.7in} 
        \captionof{table}{The proportion of each option in each question in the artist comparison User Study.}
        \vspace{-0in}
        \resizebox{\linewidth}{!}{%
        \begin{tabular}{c||cc}
        \Xhline{1.2pt}
        \diagbox{prompt-pairs}{option}  & A(our BASS) & B(artist)  \\
        \hline
        frog and broccoli       & 65.38\%  & 34.62\% \\
        eagle and cauliflower   & 68.64\%  & 31.54\% \\
        owl and tiger           & 80.00\%  & 20.00\% \\
        strawberry and dinosaur & 71.54\%  & 28.46\% \\
        giraffe and snail       & 54.62\%  & 45.38\% \\
        turtle and cat          & 68.46\%  & 31.54\% \\
        pineapple and parrot    & 73.85\%  & 26.15\% \\
        flamingo and cabbage    & 74.62\%  & 25.38\% \\
        \Xhline{1.2pt}
        \end{tabular}
        }
        \label{tab:art_userstudy}
        \centering
    \end{minipage}
\end{figure}

\begin{figure}[pht]
    \centering
    \includegraphics[width=0.77\linewidth]{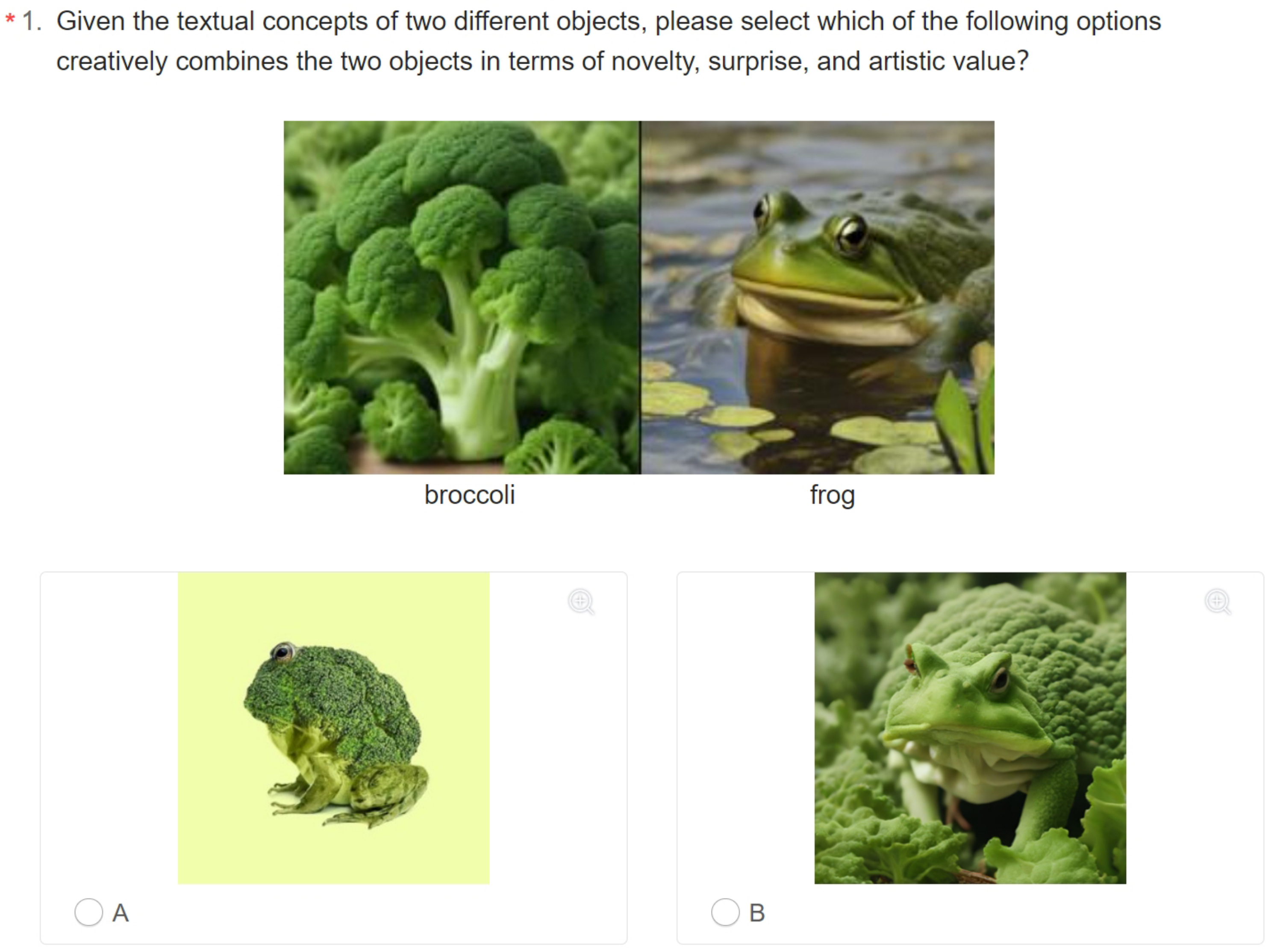}
    \caption{Interface of the artist comparison User Study.}
    \label{fig:user_study_art_inter}
    \vspace{-7pt}
\end{figure}
\begin{figure}[hpt]
    \vspace{-0pt}
    \centering
    \includegraphics[width=0.87\linewidth]{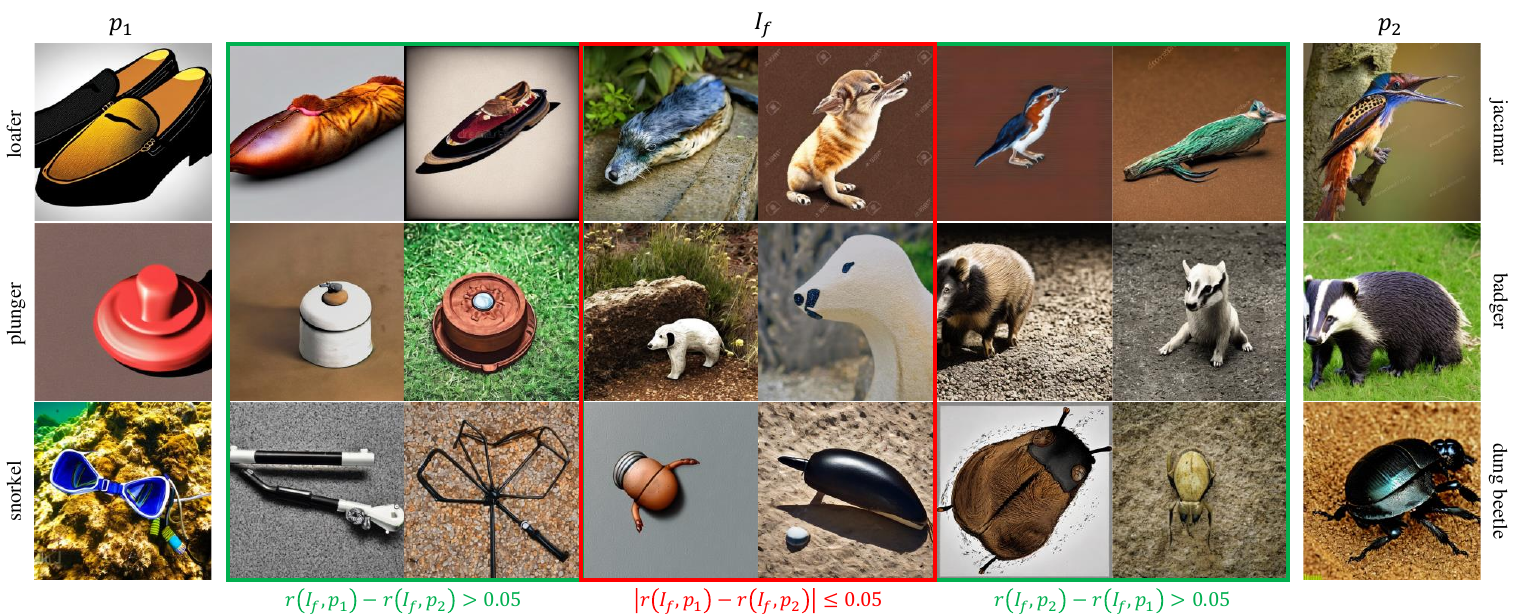}
    \caption{Visualizations with different $\theta$.}
    \label{fig:theta}
    \vspace{-7pt}
\end{figure}
\begin{figure}[pht]
    \centering
    \includegraphics[width=0.77\linewidth]{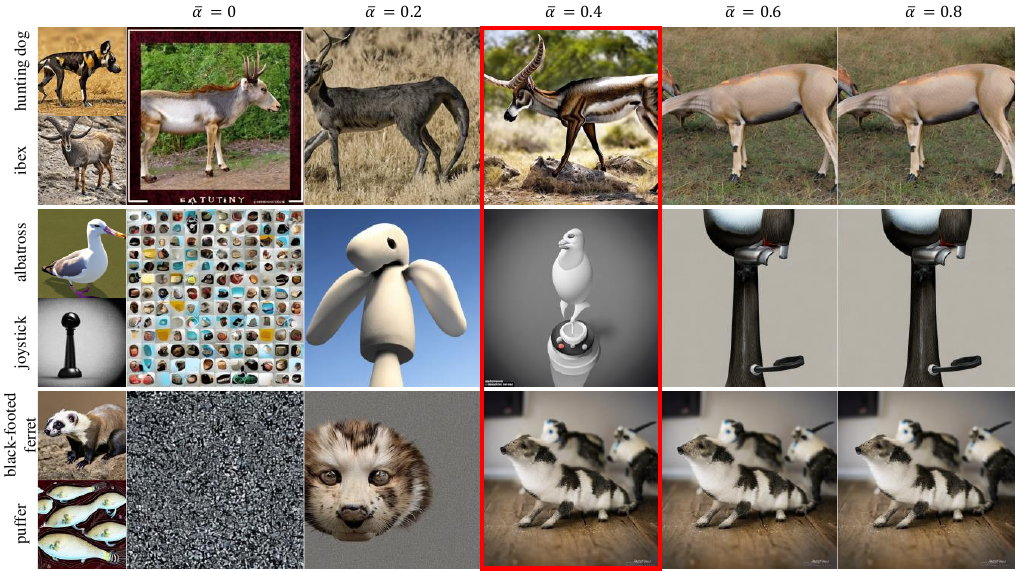}
   \vspace{-5pt}
    \caption{Visualizations with different $\overline\alpha$ when $\overline\beta=0.1$.}
    \label{fig:theta_1}
    \vspace{-7pt}
\end{figure}

\section{User Study}
\label{sec:UserStudy}
In this section, we delve into a more comprehensive explanation of our user study. In addition to the five result categories showcased in Figure \ref{fig:result_compare}, we have included an additional set of five prompt-pair groups, as illustrated in Figure \ref{fig:sup_usrstu}, to facilitate our User Study. Within these ten queries, we presented two prompts alongside their corresponding images.  The interface of the User Study is depicted in Figure \ref{fig:user_study_inter}. We gathered responses from 106 participants who diligently completed our user study, and their decisions for each subject are presented in Table \ref{tab:sup_usrstu}.

Furthermore, we present additional details of our user study comparing our results with those created by artists. We conducted the user study using 8 sets of images for comparison, with 5 sets from Figure \ref{fig:creativeobject} and the remaining 3 sets shown in Figure \ref{fig:user_stu_art}. The interface of the User Study is depicted in Figure \ref{fig:user_study_art_inter}. In this user study, we received 116 completed responses, and we display the results of each option in Table \ref{tab:art_userstudy}. From the results, it can be observed that, from the perspective of human preferences, our fused objects consistently outperform the subjectively created results of human artists.

\section{Parameter Analysis}
\label{sec:ParameterAnalysis}
\begin{figure}[ht]
    \centering
    \includegraphics[width=0.87\linewidth]{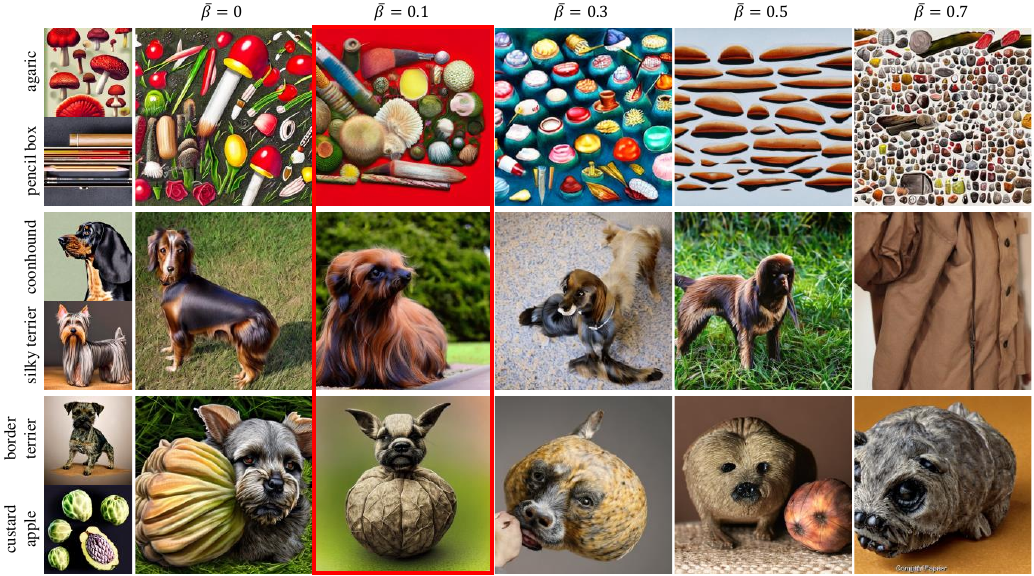}
        \vspace{-5pt}
    \caption{Visualizations with different $\overline\beta$ when $\overline\alpha=0.4$.}
    \label{fig:theta_2}
    \vspace{-5pt}
\end{figure}

Here, we showcase sampled images generated with varying values of $\theta$, $\overline\alpha$, and $\overline\beta$, as illustrated in Figures \ref{fig:theta}, \ref{fig:theta_1}, and \ref{fig:theta_2}.

In terms of the parameter $\theta$'s setting, our evaluation method is adept at selecting the most creative images from diverse distributions. This is made possible by dynamically determining $\overline\alpha$ and $\overline\beta$ based on these different distributions. However, this dynamic determination process is time-consuming. To streamline and expedite this process, we introduce a strict threshold for $\theta$ before proceeding to the overall ranking. This threshold effectively screens out images that require no further creative evaluation, as they exhibit evident biases. These biases, at this initial stage, contribute to the absence of creativity in the images, rendering them akin to straightforward outputs stemming directly from the prompt they are biased towards. We provide some illustrative results in Figure \ref{fig:theta}.

When $\overline\beta$ is set at 0.1, an increase in $\overline\alpha$ beyond our predefined value results in a biased image favoring one of the prompt pairs. Conversely, if $\overline\alpha$ falls below the predetermined threshold, the sampled image may appear unconventional due to the limited sampling space, as depicted in the shallow blue zone in Figure \ref{fig:fig_creativeimage}. This is because the optimal sample tends to be biased for acceptability. Furthermore, when $\overline\alpha$ is held constant at 0.4, varying $\overline\beta$ can help validate the assertion that lower values of $\overline\beta$ result in increased urgency, whereas higher values lead to heightened confusion.

\begin{figure}[ht]
    \centering
    \includegraphics[width=0.77\linewidth]{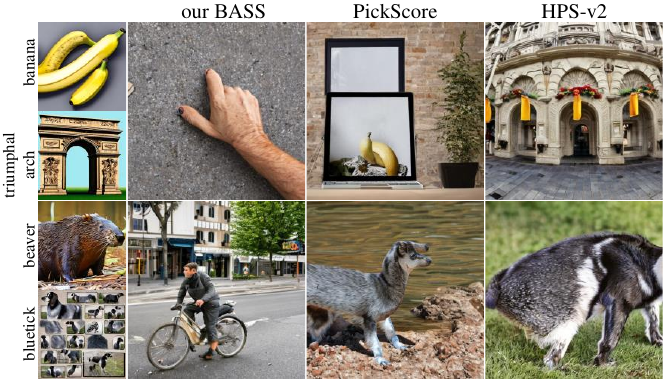}
    \caption{Failure samples in evaluation stage.}
    \vskip -0.1in
    \label{fig:fail_model}
\end{figure}
\section{Failure Examples}
\label{sec:failures}
Here, we illustrate instances of unsuccessful model outputs. Some prompt-pairs, as depicted in Figure \ref{fig:fail_model}, prove challenging to generate novel and imaginative content. Even PickScore and HPS-v2 fail to produce satisfactory samples in such cases. While increasing the sample size may potentially address this issue, the likelihood of encountering this situation is exceptionally low, estimated at approximately 5\%. Given our overall computing resources, we will refrain from excessive processing of these samples.

During the sampling stage, the primary reason for subpar samples is the inadequate configuration of hyperparameters $\alpha$ and $\beta$. Despite these values being established through consensus in our experiments, they tend to align with the distribution of majority classes, neglecting the minority classes. In our upcoming research, our focus will shift towards tailoring the distribution to the most creative samples across all categories and enhancing our model to reduce the occurrence of failed samples.

\section{Generalizations with more than two concepts.}
\label{sec:moreconcepts}
\begin{figure}[h!]
    \centering
    \includegraphics[width=0.9\linewidth]{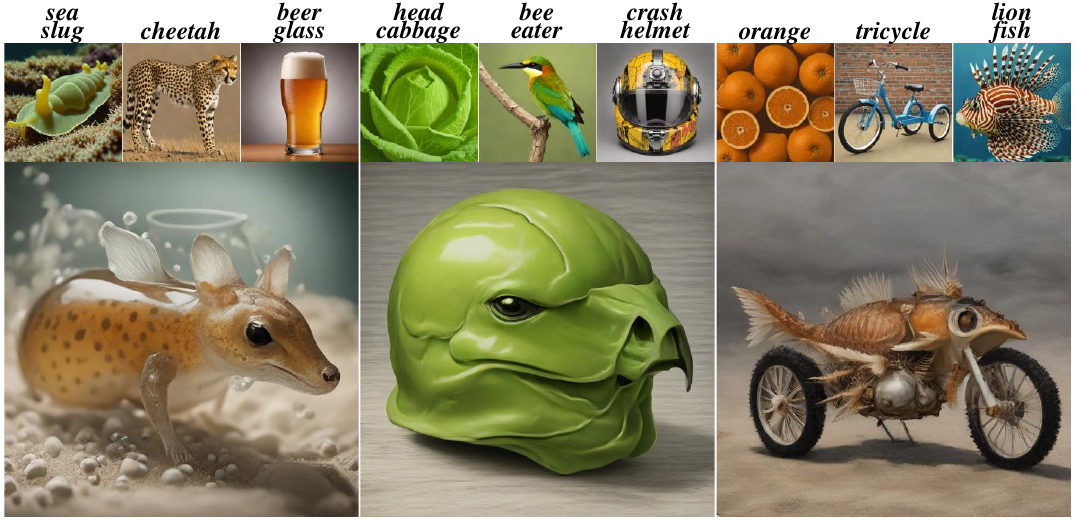}
    \vskip-0.1in
    \caption{Generalizations using more than two concepts.}
    \label{fig:result_three}
\end{figure}

We also apply our method within more than two concepts. As shown in the Figure \ref{fig:result_three}, specifically, we use our model twice to generate object images in three concepts.

\section{Comparisons with more intricate prompts}
\label{sec:intricateprompts}
\begin{figure}[h!]
    \centering
    \includegraphics[width=0.9\linewidth]{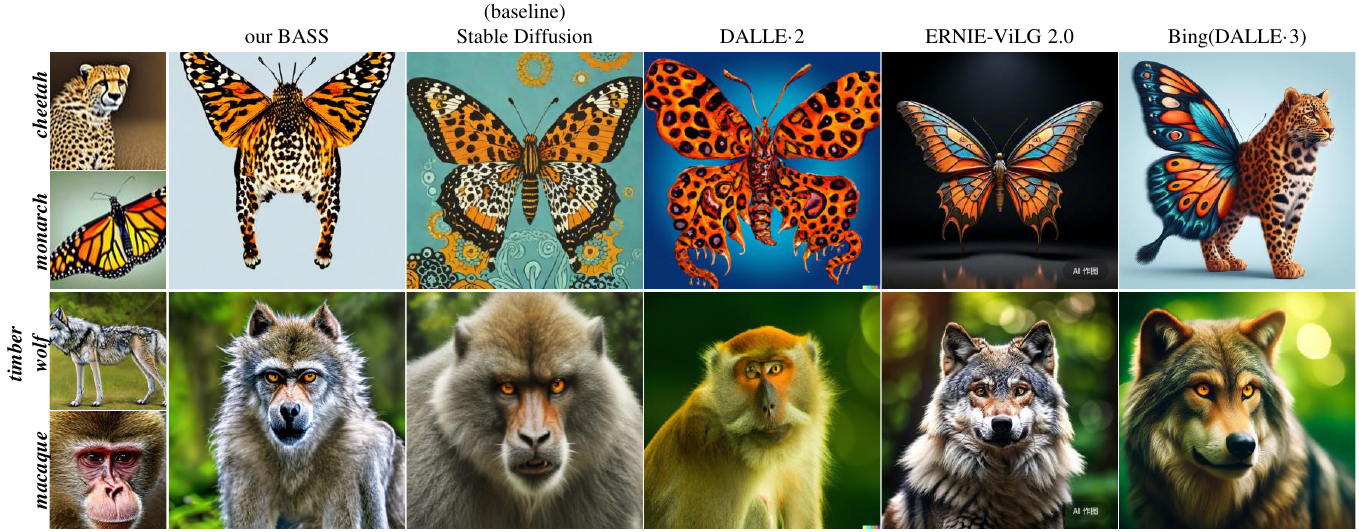}
    \vskip-0.1in
    \caption{Generalizations using more intricate prompts.}
    \label{fig:result_complex}
\end{figure}

Additionally, we incorporate enhanced prompts, such as \textit{`A monarch butterfly with a cheetah-print body and leg'}, when evaluating the compared methods in Figure \ref{fig:result_compare}.
Regrettably, their results are still lacking in creativity. Here, using more intricate prompts, such as \textit{`A surreal artwork combining elements of a butterfly and a cheetah. The creature has the upper body and wings of a vibrantly colored butterfly, featuring intricate orange and black patterns, while its lower body mimics the spotted fur of a cheetah. This imaginative hybrid stands against a soft blue background, emphasizing the striking contrast between the vivid colors of the butterfly wings and the natural tones of the leopard's fur'}, the compared methods fail to produce creative combinatorial objects as most of them only generate the different butterflies, as illustrated in Figure \ref{fig:result_complex}. DALLE-3's direct pairing of butterfly and cheetah underscores the difficulty of generating creative results solely through prompts.

\section{More Visual Results}
\label{sec:moreresults}
\begin{figure}[h]
    \vspace{-20pt}
    \centering
    \includegraphics[width=0.97\linewidth]{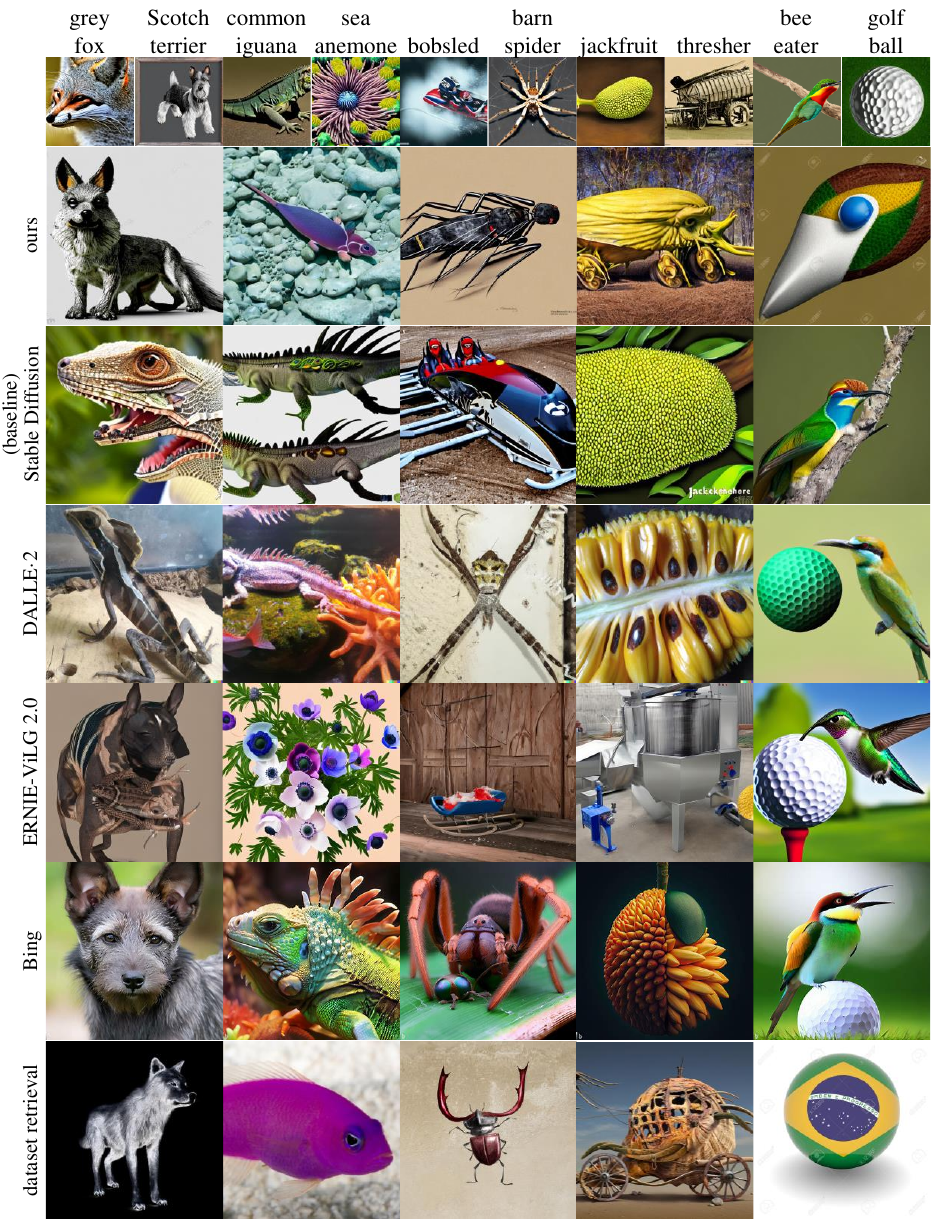}
    \caption{More visual results and comparison to other T2I models using the prompts: "Hybird of [prompt1] and [prompt2]".}
    \label{fig:more_res_cmp}
    \vspace{-15pt}
\end{figure}

\begin{figure}[h]
    \vspace{-20pt}
    \centering
    \includegraphics[width=0.97\linewidth]{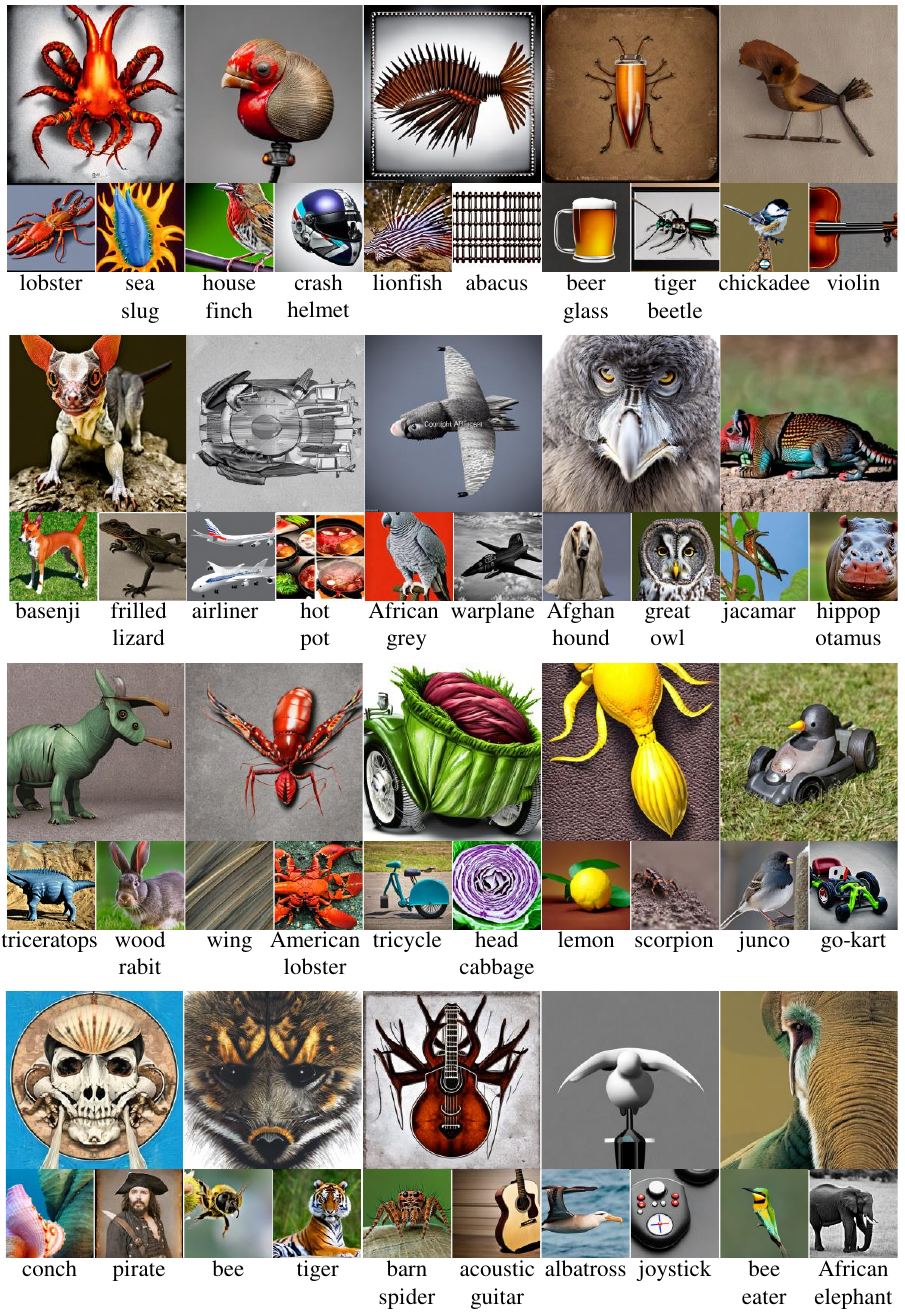}
    \caption{More visual results.}
    \label{fig:more_res_all}
    \vspace{-10pt}
\end{figure}

In the following section, we delve into supplementary findings. Figure \ref{fig:more_res_cmp} illustrates a comprehensive comparison of our results with those of other T2I models. Additionally, Figure \ref{fig:more_res_all} showcases a collection of visually captivating and groundbreaking results.

\end{document}